\documentclass[journal]{IEEEtran}

\usepackage{amsmath,amsfonts}
\usepackage{algorithmic}
\usepackage{algorithm}

\usepackage{hyperref}
\usepackage{array}
\usepackage[caption=false,font=normalsize,labelfont=sf,textfont=sf]{subfig}
\usepackage{textcomp}
\usepackage{stfloats}
\usepackage{url}
\usepackage{verbatim}
\usepackage{graphicx}
\usepackage{cite}
\usepackage{color}
\usepackage{booktabs}
\usepackage{multirow}
\begin{document}

\title{HRTransNet: HRFormer-Driven Two-Modality Salient Object Detection}
\author{Bin Tang, Zhengyi Liu*, Yacheng Tan, and Qian He
\thanks{This work is supported by the Natural Science Foundation of Anhui Province (1908085MF182), the Science Research Project for Graduate Student of Anhui Provincial Education Department (YJS20210047), and the Talent Research Fund Project of Hefei University(21-22RC14)(Corresponding author: Zhengyi Liu).}
\thanks{Bin Tang is with School of Artificial Intelligence and Big Data, Hefei University, Hefei, China(e-mail: 424539820@qq.com).}%Anhui Provincial Key Laboratory of Multimodal Cognitive Computation, Anhui University,
\thanks{Zhengyi Liu, Yacheng Tan and Qian He are with Key Laboratory of Intelligent Computing and Signal Processing of Ministry of Education, School of Computer Science and Technology, Anhui University, Hefei, China(e-mail: liuzywen@ahu.edu.cn,1084043983@qq.com,1819469871@qq.com).}
\thanks{Copyright $\copyright$2022 IEEE. Personal use of this material is permitted. However, permission to use this material for any other purposes must be obtained from the IEEE by sending an email to pubs-permissions@ieee.org.}}

\markboth{Journal of \LaTeX\ Class Files,~Vol.~14, No.~8,  February~2022}%
{Shell \MakeLowercase{\textit{et al.}}: A Sample Article Using IEEEtran.cls for IEEE Journals}

\IEEEpubid{0000--0000/00\$00.00~\copyright~2022 IEEE}

\maketitle

\begin{abstract}
The High-Resolution Transformer (HRFormer) can maintain high-resolution representation and share global receptive fields. It is friendly towards salient object detection (SOD) in which the input and output have the same resolution. However, two critical problems need to be solved for two-modality SOD. One problem is two-modality fusion. The other problem is the HRFormer output's fusion. To address the first problem, a supplementary modality is injected into the primary modality by using global optimization and an attention mechanism to select and purify the modality at the input level. To solve the second problem, a dual-direction short connection fusion module is used to optimize the output features of HRFormer, thereby enhancing the detailed representation of objects at the output level. The proposed model, named \textbf{HRTransNet}, first introduces an auxiliary stream for feature extraction of supplementary modality. Then, features are injected into the primary modality at the beginning of each multi-resolution branch. Next, HRFormer is applied to achieve forwarding propagation. Finally, all the output features with different resolutions are aggregated by intra-feature and inter-feature interactive transformers. Application of the proposed model results in impressive improvement for driving two-modality SOD tasks, e.g., RGB-D, RGB-T, and light field SOD.\url{https://github.com/liuzywen/HRTransNet}
\end{abstract}


\section{Introduction}
Salient object detection (SOD) is an important computer vision task that automatically identifies the most attractive and conspicuous objects in a scene.
SOD plays a fundamental role in image
segmentation~\cite{yarlagadda2021saliency,huang2021graph}, tracking~\cite{ma2017saliency,zhang2020non}, cropping~\cite{wang2018deep,xu2021saliency}, retargeting\cite{ahmadi2021context}, activity prediction\cite{weng2021human}, etc.%, anti-spoofing detection\cite{wang2021silicone} etc.

It continues to be challenging to apply SOD in the primary modality ``RGB image" under some conditions (e.g., poor illumination, complex backgrounds, and indistinguishable objects).
RGB-Depth (RGB-D) SOD\cite{zhou2021rgb}, RGB-Thermal (RGB-T) SOD, light field (LF) SOD\cite{fu2020light} have gradually become cutting-edge research area because of the widespread use of depth sensors, infrared cameras, and Lytro cameras.
Additional information can be  obtained by using supplementary modalities:   ``depth image" can provide additional spatial structural information; ``thermal image" can capture heat information of objects; and ``focal stack" provides several focal slices at different depths.
Effective fusion of color and supplementary modalities is critical for saliency detection.
However, most studies to date have focused on a specific type of two-modality SOD.
We explore a two-modality SOD model that is effective for  RGB-D, RGB-T, and LF SOD.

The convolutional neural network (CNN) serves as a critical backbone feature extractor  for a long time.
The CNN takes as input an image  and applies  a stack of convolution and pooling layers to expand the receptive field  and obtain a global view in deep layers.
The CNN is friendly to image classification tasks, but not dense prediction tasks, which require a decoder to gradually recover the resolution of features.

\begin{figure}[htp!]
	\centering
\begin{tabular}{cc}
  \includegraphics[width=0.23\textwidth]{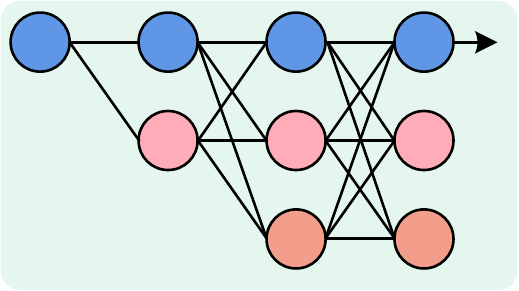}&
\includegraphics[width=0.23\textwidth]{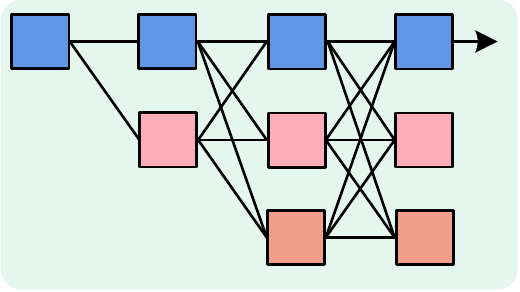}\\
\small(a) HRNet &\small(b) Revision in Convolution Layers\\
	\includegraphics[width=0.23\textwidth]{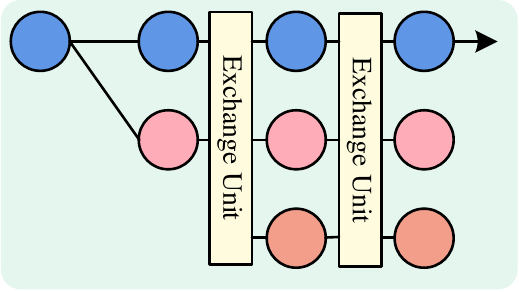}&
\includegraphics[width=0.23\textwidth]{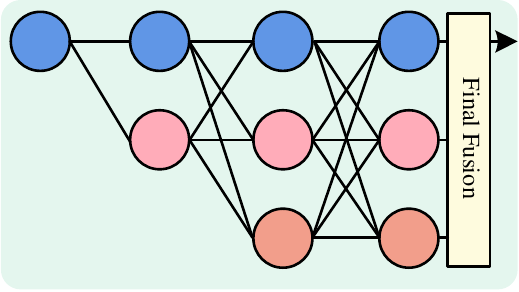}\\
\small(c) Revision in Exchange Units &\small(d) Revision in Final Fusion
\end{tabular}
	
	\caption{HRNet (a) is improved in stacked convolution layers (b), exchange units (c), and final multi-resolution fusion (d).}
     \label{fig:RevisionForHRNet}
\end{figure}

\IEEEpubidadjcol
HRNet\cite{sun2019deep} (Fig. \ref{fig:RevisionForHRNet} (a)) maintains high-resolution representation throughout the network. %, which is more adapt to dense prediction task.
%HRNet existing problem
However, the expressivity of HRNet is limited by small receptive fields and strong inductive bias from cascaded convolution operations\cite{gu2021hrvit}.
Therefore, optimized HRNets have been developed with  improved   stacked convolution layers (Fig. \ref{fig:RevisionForHRNet} (b)), the exchange units (Fig. \ref{fig:RevisionForHRNet} (c)), and final multi-resolution fusion (Fig. \ref{fig:RevisionForHRNet} (d)). The recently proposed HRFormer\cite{yuan2021hrformer} replaces convolution layer (shown as a round unit) with transformer blocks (shown as a rectangle unit) shown in Fig. \ref{fig:RevisionForHRNet} (b) to improve performance.
HRFormer can effectively perform dense prediction tasks by injecting global ability  when maintaining local features.

In this study, we use HRFormer as a backbone network to drive two-modality SOD tasks, in particular, RGB-D, RGB-T, and LF SOD.
Two issues need to be addressed:
(1) How can two-modality fusion be achieved using HRFormer?
(2) Can HRFormer be further improved?

Ren et al.\cite{ren2021dual} propose dual-stream HRNet to perform two-modality fusion, that is,
two types of heterogeneous data (synthetic aperture radar and optical image) are combined. Features with different modalities are fused in the high-resolution branch. This design generates an insufficient receptive field.
AVT\cite{sajid2021audio} exploits auditory information to
aid  visual models in  crowd counting tasks. Audio embedding is only integrated into image features in the last three-branch exchange unit. Audio modality is shallow-modeled, which is not suitable for supplementary image modalities, such as the depth image, thermal image or focal stack.

Unlike existing fusion methods,  an auxiliary stream is used in this study to encode a supplementary modality and inject the modality into HRFormer at the beginning of each multi-resolution branch. The supplementary modality is deep-modeled to exploit abundant information and enhance representation.

\begin{figure}[htp!]
	\centering
\begin{tabular}{cc}
  \includegraphics[width=0.097\textwidth]{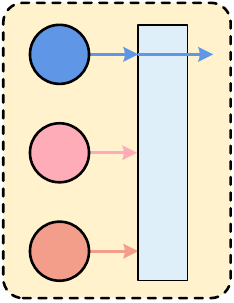}&
\includegraphics[width=0.101\textwidth]{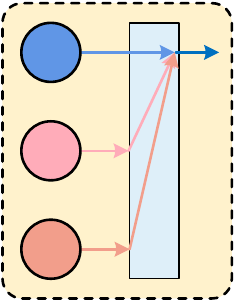}\\
\small(a) HRNetV1&\small(b) HRNetV2\\
\end{tabular}
\begin{tabular}{ccc}
	\includegraphics[width=0.1\textwidth]{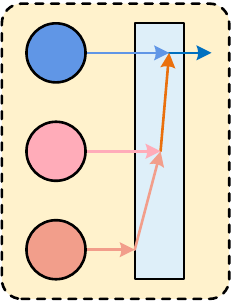}&
\includegraphics[width=0.1\textwidth]{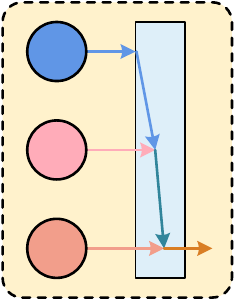}&
\includegraphics[width=0.1\textwidth]{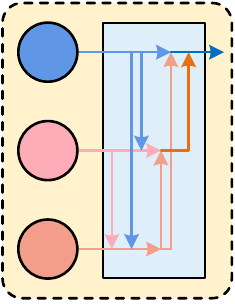}\\
\small(c) HRNetV2p-V1&\small(d) HRNetV2p-V2&\small(e) Ours
\end{tabular}
	\caption{Comparison of final multi-resolution fusion methods among HRNetV1, HRNetV2, HRNetV2p-V1, HRNetV2p-V2, and ours.\label{fig:RevisionOutput}}
\end{figure}

Fig. \ref{fig:RevisionOutput} shows the final multi-resolution fusion:
HRNetV1\cite{sun2019deep} only outputs a high-resolution feature map for pose estimation;
HRNetV2\cite{sun2019high,wang2021deep} concatenates all the output features for semantic segmentation and facial-landmark detection;
HRNetV2p-V1 and HRNetV2p-V2\cite{sun2019high,wang2021deep} output feature pyramid representation from high-to-low or from low-to-high for object detection and classification tasks.

Unlike the aforementioned networks, a dual-direction short connection fusion module  is designed in this study, that consists of four \textbf{I}ntra-feature and \textbf{I}nter-feature \textbf{I}nteractive \textbf{T}ransformers (TripleITs).
The proposed module is used to  optimize each resolution output feature by itself and the other resolution features, to accurately depict the details of an object.

The four main contributions of this study are summarized below:
\begin{itemize}
    \item
    A unified two-modality SOD model (HRTransNet) is proposed for RGB-D, RGB-T, and LF SOD tasks. This model can  be further applied to other dense prediction tasks with two modalities. HRTransNet is built upon the HRFormer backbone and maintains high-resolution representation with a large receptive field achieved by transform blocks.
    \item A  module is proposed to inject a supplementary modality into the primary modality by formulating the weight of each supplementary modality from a global perspective and emphasizing the coordinate-wise spatial position of the supplementary modality, thereby achieving two-modality fusion at  the input level.
    \item A novel designed dual-direction short connection fusion module is used to optimize a resolution feature by itself and the other resolution features, thereby improving the final multi-resolution fusion of HRFormer at the output level.
     \item    The proposed model exhibits excellent performance for the  RGB-D, RGB-T, and LF datasets, which demonstrates the  superiority of the model for two-modality complementary tasks.
\end{itemize}

\section{Related works}
\subsection{High-resolution network}
HRNet\cite{sun2019deep} maintains high-resolution representations in the forwarding propagation process by generating feature maps with different resolutions in parallel and repeatedly conducting multi-scale fusions in the exchange unit, which is friendly to dense prediction tasks.
%introduce exchange units across parallel subnetworks such that each subnetwork repeatedly receives the information from other parallel subnetworks.
%HRNet maintains high-resolution representations through the whole process by connecting high-to-low resolution features in parallel and repeatedly conducts multi-scale fusions across parallel features.
%HRNet retains the higher representations and performs high-low convolutions in parallel and in multiple streams, ensuring spatially precise learning.
%HRNet develops a creative method to maintain the high-resolution representation of input through the whole inference process of model.
%HRNet connects high-to-low resolution convolutions in parallel and repeatedly implements fusions across parallel convolutions for high-resolution representations.
HRNet has been widely applied for human-pose estimation\cite{sun2019deep,zhang2021human}, semantic segmentation\cite{wu2021optimized}, facial-landmark detection\cite{sun2019high}, surface-defect detection\cite{akhyar2021beneficial}, video tracking\cite{shao2021hrsiam}, image inpainting\cite{wang2021parallel}, remote-sensing pansharpening\cite{wu2021dynamic}, and gaze estimation\cite{cai2021gaze}.

A few methods  have recently been developed to optimize HRNet.
Improvements are made to the stacked convolution layers, exchange unit and final multi-resolution fusion, shown in Fig. \ref{fig:RevisionForHRNet}.

%convolution layer
The following modifications to stacked convolution layers  have been proposed: in
Lite-HRNet\cite{yu2021lite}, a conditional channel weight block  replaces the convolution layer.
Zhang et al.\cite{zhang2021human} add an attention module  before the convolution layers of the last stage.
Wang et al.\cite{wang2021parallel} use
four resolutions at the beginning rather than adding resolution
gradually, and apply a shared attention map to all the resolution features of the last stage.
In HRFormer\cite{yuan2021hrformer}, HRViT\cite{gu2021hrvit}, and HR-NAS\cite{ding2021hr}  the convolution layer is replaced with a transformer block\cite{liu2021swinnet,dong2021cswin} to expand the receptive field.
%HRViT\cite{gu2021hrvit} constructs a multi-scale high-resolution ViT backbone by jointly optimizing key building blocks with efficient embedding layers, augmented cross-shaped attentions, and mixed-scale convolutional  feed-forward networks.
%HR-NAS\cite{ding2021hr} inserted a lightweight transformer path into the residual blocks to extract global information and applied the neural architecture search to remove the channel/head redundancies.

%exchange unit
The following modifications to the exchange unit have been proposed:
Yang et al.\cite{yang2021scale} use an attentive multi-scale fusion  module to modify  feature fusion in  the exchange units. This scheme is  flexible  for different types of information and lets the network focus on more discriminate features.
%exchange unit
Zhao et al.\cite{zhao2021human} replace the original exchange unit with a gated multi-scale  fusion module to eliminate  noise ambiguities.
In Lite-HRNet\cite{yu2021lite} and HRViT\cite{gu2021hrvit}, normal convolutions are replaced with depthwise separable  convolutions\cite{chollet2017xception} in the exchange unit.
In AVT\cite{sajid2021audio},  a transformer-inspired attention mechanism is deployed to perform inter-branch fusion.
%AVT\cite{sajid2021audio} proposes Visual Feature Extraction network which is similar with HRNet. It comprises of three multi-scale branches with repeated interbranch fusion which deploys the transformer-inspired attention mechanism.

%final fusion
Fig. \ref{fig:RevisionOutput}  shows the modifications proposed for final multi-resolution fusion:
HRNetV1\cite{sun2019deep} only outputs the high-resolution feature map.
HRNetV2\cite{sun2019high,wang2021deep} concatenates all the output features.
HRNetV2p\cite{sun2019high,wang2021deep} outputs the feature pyramid representation from high-to-low or from low-to-high.
%final fusion
Ding et al.\cite{ding2021learning} introduce a soft conditional
gate module\cite{li2020learning} at the last stage to generate dynamic weights reflecting the importance of different features for fusion.
%final fusion
Wu et al.\cite{wu2021optimized} use mixed dilated convolution to enhance the scale-awareness ability of output features and   data-dependent upsampling\cite{tian2019decoders} to progressively fuse multi-level features.
%final fusion
Yang et al.\cite{yang2021scale} gradually combine  low-to-high resolution features by interpolation, deconvolution,  and a series of convolutions to extract the local and global features involved at all resolutions.
%final fusion
Yang et al.\cite{yang2021dense} aggregate the feature maps of different layers using spatial attention and channel attention.

In this study, we drive the model using HRFormer  in which the stacked convolution layers have been modified. Furthermore, we introduce a transformer-based fusion strategy for final multi-resolution fusion.
This strategy effectively exploits the correlation between intra- and inter-features.

In addition,
in the multi-modality fusion area,
Ren et al.\cite{ren2021dual} propose a dual-stream HRNet to combine two types of heterogeneous data. Features with different modalities are fused in   high-resolution branches. An insufficient receptive field is generated.
AVT\cite{sajid2021audio} embeds the audio modality into the image modality only in the last three-branch exchange unit.

Unlike existing fusion methods, in this study, a supplementary modality encoded by the auxiliary stream is injected into HRFormer at the beginning of each multi-resolution branch.
\subsection{Two-modality salient object detection}
Initially, we detect the salient objects in a color image using conventional methods\cite{cong2018review,liu2018salient,zhou2019adaptive} or CNN-based\cite{wang2020deep,zhu2019aggregating,hu2020sac,cong2021rrnet,fang2021densely,zhang2019synthesizing} methods. %ren2018multi,xu2020dual,fang2021ibnet,
However, some objects are very difficult to detect, irrespective of the method used.
This result may be obtained because of complex backgrounds or indistinguishable objects in a scene.
Additional
information that is not well represented in a color image can be obtained using supplementary modalities, e.g., the depth image (D),
thermal image (T), and focal stack (FS), using widely used depth sensors, infrared imaging devices, and Lytro cameras.
For example, the depth image is effective for obtaining the spatial information, %thermal modality has advantage in weak light condition,
the thermal image can capture the heat radiated from objects under poor illumination, and the focal stack provides focus cues at different depths.
%which are synthesized through light field rendering and refocusing techniques provide focus cues and a more comprehensive natural scene information, also including depth information.
Therefore, the depth and thermal image and the focal stacks serve as supplementary modalities for  improving SOD.

RGB-D, RGB-T and LF (RGB-FS) SOD all use two modality fusion to comprehensively segment objects.
Early fusion\cite{liu2019salient}, middle fusion\cite{wen2021dynamic,zhu2021rgb,zhang2020revisiting,li2020asif} and late fusion~\cite{wang2022learning,liu2020cross,shu2022expansion} are three classic frameworks.
The performance of  RGB-D SOD can be improved using an attention mechanism\cite{liu2020learning}, edge guidance\cite{jiang2020cmsalgan,liu2021cross}, depth calibration\cite{ji2021calibrated}, depth estimation\cite{zhao2021rgb,wu2021modality}, depth quality assessments\cite{fan2020rethinking,wang2021depth}, 3D convolution\cite{chen2021rd3d},  deformable convolution\cite{li2021amdfnet}, automatic architecture search\cite{sun2021deep},
uncertainty distribution\cite{zhang2020uc}, and bi-directional guidance\cite{yang2022bi,zhang2021cross}.
To reduce the influence of poor depth images, Cong et al.\cite{cong2017iterative} use salient seed diffusion with a depth constraint and introduce a  depth confidence weight\cite{cong2016saliency}.
Chen et al.\cite{chen2020dpanet} learn the potential confidence of depth map that will be aggregated into multi-modality fusion.

RGB-D SOD  is studied prevalently, whereas, RGB-T and LF SOD are  less exploited.
%The thermal image can capture the radiated heat of objects in worse illumination. It serves as the supplementary to color modality.
The support vector machine\cite{ma2017learning}, ranking algorithm\cite{tu2019m3s,wang2018rgb, tang2019rgbt}, and graph learning\cite{tu2019rgb} were initially used for RGB-T SOD. Later, CNN based methods (e.g., ECFFNet\cite{zhou2021ecffnet}, MIDD\cite{tu2021multiinteractive}, MMNet\cite{gao2021unified}, CGFNet\cite{wang2021cgfnet}, CSRNet\cite{huo2021efficient}, CGMDRNet\cite{chen2022cgmdrnet}, and SwinNet\cite{liu2021swinnet}) were used to  exploit two-modality fusion within attention, boundary, local and global contexts, depth-guidance, and cross-guidance perspectives.
%Light field SOD
%Light field inherently captures a structured 4-D representation.
The focal stack indicates different focused regions in different depth layers for LF SOD.
Early methods \cite{li2014saliency} and \cite{li2015weighted} introduced prior knowledge and  weighted sparse coding. Later, ConvLSTM  was employed\cite{piao2020exploit,wang2019deep,zhang2020lfnet,zhang2019memory}  to  process the focal stack due to sequential attribution. Recently widely used techniques (e.g., 3D convolution\cite{zhang2021SANet} and graph  networks\cite{liu2021light}) have been introduced to model  information fusion within the focal stack.

The summary on RGB-D, RGB-T, and LF SOD methods, presented above shows that each of these methods can be used to perform one or two tasks\cite{gao2021unified,liu2021swinnet}. Few models have been designed to perform two-modality SOD.
The three tasks exhibit  common characteristics.
A supplementary modality needs to be purified in  combination with primary modality because of the presence of noise.
Irrespective of the supplementary
modality is used, SOD remains the dense prediction task that must to be performed to maintain the resolution between the input and output.
Thus, we adapt HRFormer to a two-modality SOD task.

We also identify  other two-modality SOD tasks (e.g., image-video optical flow  SOD\cite{yang2021learning}, text-image SOD\cite{li2021joint}, audio-image  SOD\cite{chen2021audiovisual}) and the similar but different image pair co-saliency task\cite{zhang2016co,zhang2021summarize,cong2022global}. %{\color{blue}and activity recognition (RGB image and skeleton)\cite{shu2022expansion}}.
Therefore, two-modality SOD is worthy of further study.

\subsection{Transformer-based salient object detection}
In the past year, transformers have been successfully employed to execute computer vision tasks.
GLSTR\cite{ren2021unifying} is the first case in which a pure transformer-based encoder has been applied for SOD.
%SOD transformer
%Unifying Global-Local Representations in Salient Object Detection with Transformer
Transformers were subsequently gradually applied  in research on RGB-D SOD\cite{zhou2021rgb} and co-saliency\cite{tang2021cosformer}.
RGB-D SOD has been achieved using
%TriTransNet
TriTransNet\cite{liu2021tritransnet}, where three transformers with shared weights are employed to enhance the representation ability of high-layer features.
%Luhuchuan-TransCMD
TransCMD\cite{pang2021transcmd} decodes multi-scale and multi-modal features using a transformer. These features are progressively integrated by self-attention and cross-attention among different modalities and scales.
%TransCMD\cite{pang2021transcmd} fuses color and depth feature in the same layer with intra-modal self-attention and inter-modal cross-attention, and further enhances the decoded feature by cross-scale self-attention in the decoding process.
%Luhuchuan-Transformer-based Network for RGB-D Saliency Detection
Wang et al.\cite{wang2021transformer} realized the feature enhancement and fusion using a transformer decoder. Multi-head self-attention is used to refine the initial fused feature and enhance the feature at every scale.
%The initial fused feature is refined by all the multi-scale and multi-modal features to achieve the feature fusion, and meanwhile the feature on each scale is enhanced by integrating complementary features of all other scales.
%Wang et al.\cite{wang2021transformer} apply the structure of transformer decoder to automatically fuse multi-modal and multi-scale features.
%JiangBo-MutualFormer
MTFNet\cite{wang2021mtfnet} learns intra-modality and inter-modality communication simultaneously by using a self-attention and cross-attention transformer in the encoder section.

%Transformer backbone
Transformer-based backbones have exhibited superior performance.
%VST
The Visual Saliency Transformer (VST)\cite{liu2021visual} divides an image into patches and uses T2T-ViT\cite{yuan2021tokens} to  propagate long-range dependencies between image patches. VST also designs a reverse T2T decoder simultaneously. %and introduces edge detection to improve the performance.
%Transformer Transforms Salient Object Detection and Camouflaged Object Detection
In TransformerSOD\cite{mao2021transformer},  a swin transformer is used as the backbone, and a generative adversarial network and difficulty-aware learning are used to produce saliency predictions.
%the generator to produce saliency predictions with deep supervision and the confidence estimation based discriminator to estimate pixel-wise confidence map and achieve difficulty-aware learning.
%SwinNet
SwinNet\cite{liu2021swinnet} also uses a Swin transformer as the backbone, which is used in combination with edge information to perform  decoding.
%ZhangJing-TEP
Zhang et al.\cite{zhang2021learning} use maximum likelihood estimation to train a generative vision transformer network.
%propose a generative vision transformer network by an energy-based informative prior distribution defined on latent space for salient object detection. They jointly train the vision transformer network and the energy-based prior model by maximum likelihood estimation, without relying on any extra assisting network for adversarial learning or variational learning.
\begin{figure*}[t]
	\centering
	\includegraphics[width=1\linewidth]{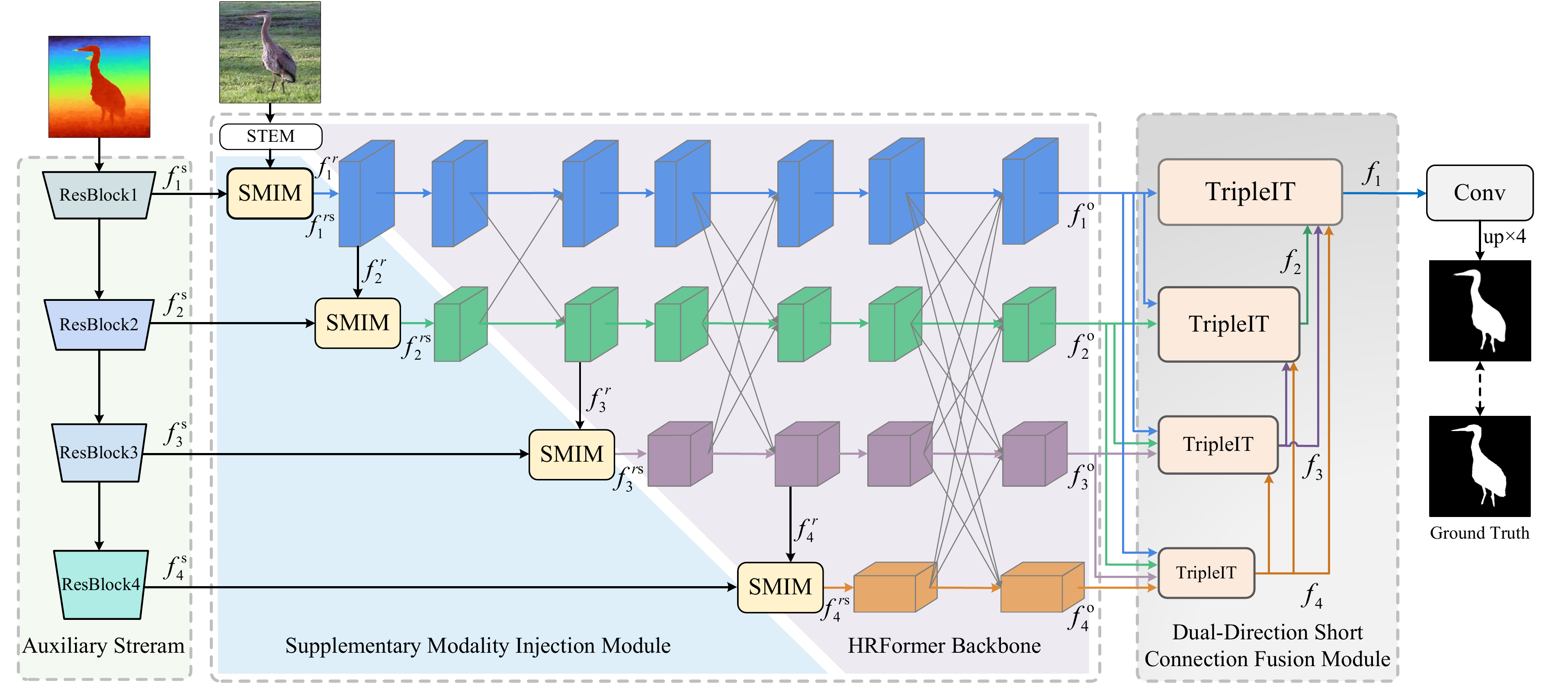}
	\caption{Take as an example RGB-D salient object detection to show the network architecture of the proposed HRTransNet which is also applied in RGB-T and light field salient object detection.
It consists of auxiliary stream, supplementary modality injection module, HRFormer backbone, and dual-direction short connection fusion module. } \label{fig:main}
\end{figure*}

Unlike  existing methods,  HRFormer with high-resolution representation is used in this study to achieve two-modality SOD. The advantages of both the high-resolution network and the transformer are fully exploited in this effective combination.

\section{Proposed method}

\subsection{Motivation}
The high-resolution network (e.g., HRNet and HRFormer) maintains high-resolution representation throughout forwarding  propagation, which is friendly to SOD task in which the input and output have the same resolution.
Therefore, we attempt to apply HRFormer to SOD.
However, two-modality SOD involves the fusion of two modalities.
The  supplementary modality of the ``depth image", ``thermal image" and  ``focal stack" can provide the primary modality of the ``color image" with spatial, thermal infrared, and focus information, but have some defects (e.g., noise, local blurring, and redundancy).
Thus, we design a module that injects supplementary modalities into the primary color modality by formulating the weight for each supplementary modality from a global perspective and emphasizing the coordinate-wise spatial position of the supplementary modality.
In addition, HRFormer improves upon HRNet by replacing convolution blocks with transformer blocks without changing the exchange unit and final multi-resolution fusion.
However, multi-scale information is vital in two-modality SOD for capturing the  details of an object. Consequently,  we design another important module that has four transformer-like fusion units with dual short connections. Each transformer-like fusion unit is called an intra-feature and inter-feature interactive transformer (TripleIT). Each TripleIT takes an output resolution feature of HRFormer as the primary feature, takes higher resolution output features  of HRFormer and lower resolution output features of TripleIT as associate features, and models intra- and inter- feature correlation. The designed module enhances multi-scale feature fusion  and thereby the details of salient objects.
In Fig. \ref{fig:main},  RGB-D SOD is used  as an example  to illustrate the network architecture of HRTransNet, which can also be applied to RGB-T and LF SOD. Note that the depth map adopts Turbo colormap\footnote{\href{https://ai.googleblog.com/2019/08/turbo-improved-rainbow-colormap-for.html}{https://ai.googleblog.com/2019/08/turbo-improved-rainbow-colormap-for.html}} for the better visualization.
The model comprises four components, an  auxiliary stream, a supplementary modality injection module, an HRFormer backbone, and a dual-direction short connection fusion module, which are described in the following sections.

\subsection{Backbone network for primary color modality}
The HRFormer\cite{yuan2021hrformer} is an effective backbone of  the primary color modality, because it can maintain high-resolution representations throughout the network that accords with the SOD task in which the input and output have the same resolution.

Therefore, we use HRFormer to extract rich and productive visual representations.
HRFormer has a high-resolution convolution stem and a main body with three progressive high-to-low resolution stages.
%The multi-resolution streams are connected in parallel.
%The main body consists of a sequence of stages.
Each stage of the main body contains a sequence of transformer blocks and  exchange units.
The transformer blocks perform feature updates at the same resolution, and the exchange units perform information exchange across the different resolutions.
Unlike HRNet\cite{sun2019deep}, HRFormer uses transformer blocks to enlarge  receptive fields and achieve a global perspective. The exchange units in HRFormer are not modified.

For the purpose of clarity in the following discussion of fusion with a supplementary modality, we define the initial color features for the  four stages of HRFormer as $F^r=\{f^r_i\}^{4}_{i=1}$, which are shown in Fig. \ref{fig:main}.
%The stages of HRFormer includes stem, stage2, stage3, stage4.
%$f^r_1$ is the feature after the stem of HRFormer.
%$f^r_4$ is down-sampled smallest scale features after multi-resolution fusion module of
%In the input of 3,4 stage
%The smallest scale features after multi-resolution fusion module of each stage are down-sampled, and they are defined as initial color features $f^r_i$.

\subsection{Auxiliary stream for supplementary modality}
Supplementary modality in RGB-D, RGB-T, and LF SOD is depth image, thermal image and focal stacks, respectively.
%Supplementary modality can provide important complementary information for identifying salient objects. Nevertheless, the low-quality versions often cause wrong detections\cite{cong2016saliency}.
To find discriminative and complementary information to form the supplementary to primary color modality, we use an auxiliary stream to extract supplementary features.
Considering the performance and computation cost, ResNet18\cite{he2016deep} is selected as the auxiliary backbone. The detailed analysis can be seen in Section \ref{sec:auxiliarystream}.
The extracted hierarchical supplementary features are denoted as $F^s=\{f^s_i\}^{4}_{i=1}$, where $i$ denotes layer number.

\subsection{Supplementary modality injection module}
\begin{figure}[htp!]
	\centering
	\includegraphics[width=0.9\linewidth]{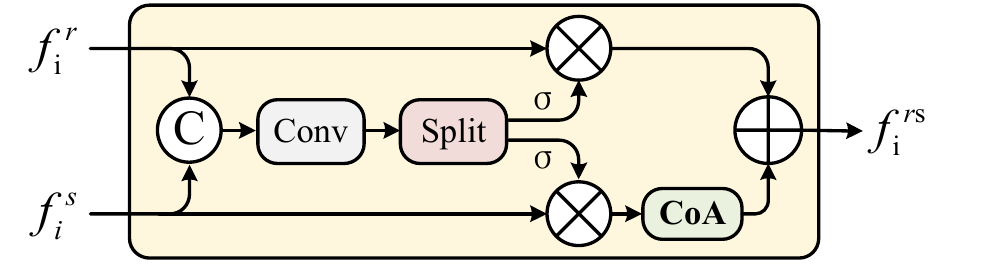}
	\caption{Supplementary modality injection module (SMIM). \label{fig:SMIM}}
\end{figure}

To use the high-resolution merit of HRFormer, supplementary modality is injected into primary color modality at the beginning of each stage. %Inspired by AMC-Net\cite{yang2021learning},
Fig. \ref{fig:SMIM} shows the detail of the proposed supplementary modality injection module (SMIM).
It achieves two-modality fusion between initial color features $f^r_i\left(i=1,\cdots,4\right)$ of primary color modality and supplementary features $f^s_i\left(i=1,\cdots,4\right)$ of another supplementary modality.

To be specific, primary color feature $f_i^r$  and supplementary features $f^s_i$ are first concatenated and convoluted to form a two-channel feature. Then it is split to represent two modalities. Next, sigmoid activation function and average pooling operation are successively applied to generate the weight of each modality. The process can be described as:
\begin{equation}
\begin{aligned}
\left[w_i^r,w_i^s\right]=\textit{Avg}\left(\textit{sigmoid}\left(\textit{split}\left(\textit{Conv}\left(\textit{Cat}\left(f_i^r,f_i^s\right)\right)\right)\right)\right)
\end{aligned}
\end{equation}
where $\textit{Cat}(\cdot)$ is concatenation operation, $\textit{Conv}(\cdot)$ is a convolution operation to generate a two-channel feature, $\textit{split}(\cdot)$ is a split operation along channel direction to yield two single-channel maps, $\textit{sigmoid}(\cdot)$ is sigmoid activation operation, and $\textit{Avg}(\cdot)$ is average pooling operation. Then two weights $w_i^r$ and $w_i^s$ are generated. They reflect the importance of each modality feature on the final result.

Next, the weights $[w_i^r,w_i^s]$  are reassigned to color and supplementary features by element-wise multiplication operation.
Especially, supplementary modality further applies a coordinate attention\cite{hou2021coordinate} to focus on more attentive region by its direction-aware and position-sensitive ability.
As a result, the noises  are suppressed in supplementary modality and features are purified more discriminative.
Finally, two modalities are combined by element-wise addition operation to produce new initial features $F^{rs}=\{f^{rs}_i\}^{4}_{i=1}$. The process can be described as:

\begin{equation}
\begin{aligned}
f_i^{rs}=w_i^r \times f_i^r +\textit{CoA}\left(w_i^s \times f_i^s\right)
\end{aligned}
\end{equation}
where ``+" is element-wise addition operation, ``$\times$" is element-wise multiplication operation, and $\textit{CoA}\left(\cdot\right)$ is coordinate attention\cite{hou2021coordinate}.

\subsection{Dual-direction short connection fusion module}
%advantage of HRFormer
HRFormer applies transformer blocks to enlarge receptive field of fused feature $F^{rs}$, and uses exchange units to absorb the merits of multi-scales features. The process is described as:
\begin{equation}
\begin{aligned}
F^{o}=\textit{HRFormer}\left(F^{rs}\right)
\end{aligned}
\end{equation}
After HRFormer, output features $F^{o}=\{f^{o}_i\}^{4}_{i=1}$ achieve information exchange and optimization in both global and local scopes.

Next, aggregating information from output features $f^{o}_i(i=1,\cdots,4)$ of HRFormer is an essential operation for dense prediction tasks. The feature concatenation dominates the choice of aggregation operations, but its expressiveness is limited.
%exist problem
Therefore, we devise an dual-direction short connection fusion module to aggregate all the multi-resolution output features of HRFormer to generate decoding features $F^{}=\{f^{}_i\}^{4}_{i=1}$. It consists of four Intra-feature and Inter-feature Interactive Transformers (TripleIT) $\mathbb{T}=\{T^{}_i\}^{4}_{i=1}$.
In the left of module, from top to bottom, high-resolution features are appended to  all the low-resolution features. For example, $f_1^o$ is added to $T_2,T_3,T_4$; $f_2^o$ is added to $T_3,T_4$; $f_3^o$ is added to $T_4$.
In the right of module, from bottom to top, decoding  features $f^{}_i$ are added to all the TripleIT $T_j(j=i-1,\cdots,1)$ with higher resolution.
For example, $f_4$ is added to $T_3,T_2,T_1$; $f_3$ is added to $T_2,T_1$; $f_2$ is added to $T_1$.
Therefore, each TripleIT $T_i$ is denoted as:
        \begin{equation}
\begin{aligned}
f_i= T_i\left(f^o_i,f^{asso}_i\right)\left(i=1,\cdots,4\right)
\end{aligned}
\end{equation}
where $\boldsymbol{f_i^o}$  serves as primary feature of $T_i$, the rest of input features serve as associated features $f^{asso}_i$ that are  defined as:
\begin{equation}
\begin{aligned}
f^{asso}_i= \left\{\begin{matrix}
    \textit{FlatCat}\left(f_2,f_3,f_4\right),&i=1\\
	\textit{FlatCat}\left(f_1^o,f_3,f_4\right),&i=2\\
	\textit{FlatCat}\left(f_1^o,f_2^o,f_4\right),&i=3 \\
	\textit{FlatCat}\left(f_1^o,f_2^o,f_3^o\right),&i=4
	\end{matrix}\right.
\end{aligned}
\end{equation}
where $\textit{FlatCat}\left(\cdot,\cdot,\cdot\right)$ flattens all the features and concatenates them in the patch direction.
%\begin{equation}
%\begin{aligned}
%f_i= \left\{\begin{matrix}
%    T_i(\boldsymbol{f_i^o},f_2,f_3,f_4),&i=1\\
%	T_i(f_1^o,\boldsymbol{f_i^o},f_3,f_4),&i=2\\
%	T_i(f_1^o,f_2^o,\boldsymbol{f_i^o},f_4),&i=3 \\
%	T_i(f_1^o,f_2^o,f_3^o,\boldsymbol{f_i^o}),&i=4
%	\end{matrix}\right.
%\end{aligned}
%\end{equation}

Each primary feature is first optimized by long-range dependency of itself, and then comprehensively combined with associated features, to generate decoding features. The details are discussed in the following section.

At last, $f_1$ is convoluted and 4$\times$upsampled to form saliency map.
%The process can be described as:
%\begin{equation}
%\begin{aligned}
%f_i= TripleIT(a,b)\\
%\end{aligned}
%\end{equation}
%where $a$ is main feature, and $b$ is associated features.
\subsection{Intra-feature and inter-feature interactive transformer}
\begin{figure}[htp!]
	\centering
	\includegraphics[width=1\linewidth]{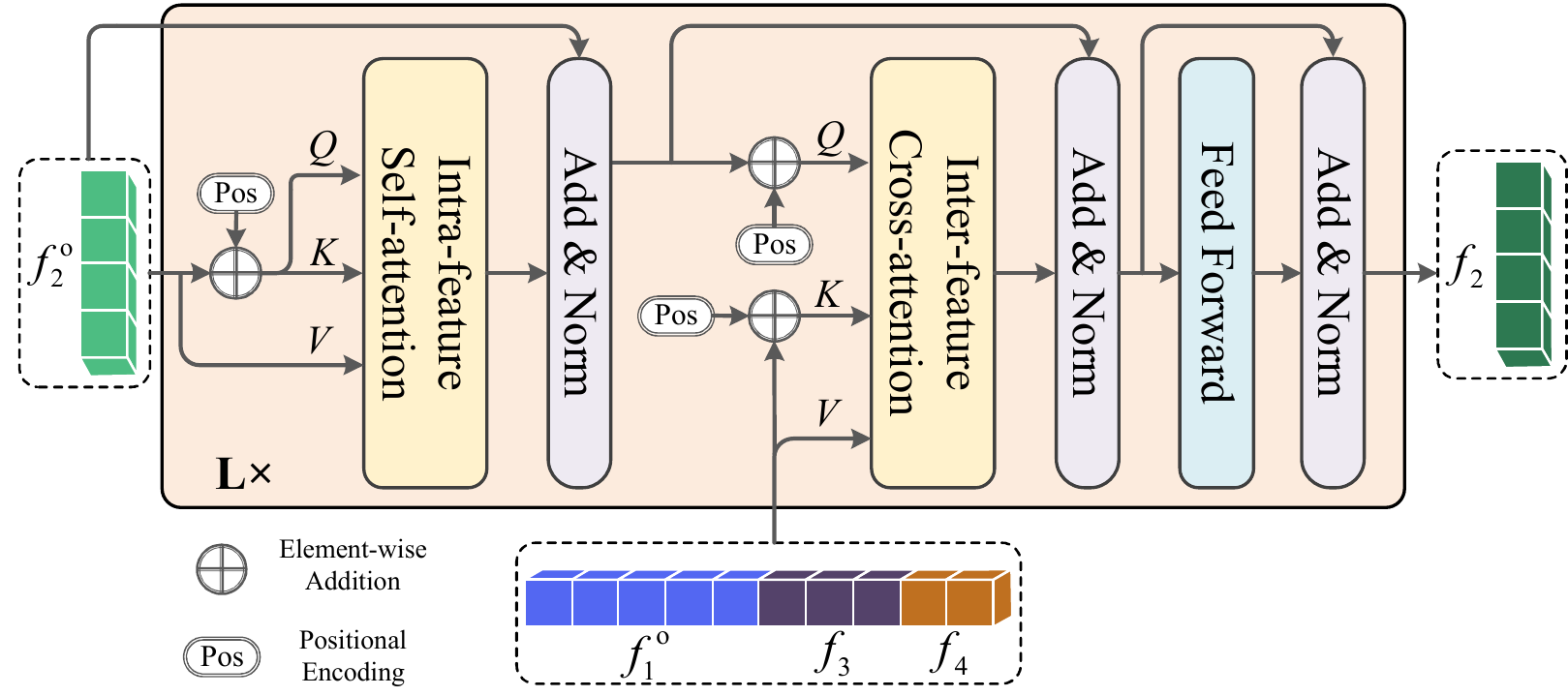}
	\caption{Intra-feature and inter-feature interactive transformer (TripleIT). \label{fig:TripleIT}}
\end{figure}

Intra-feature and inter-feature interactive transformer (TripleIT) models the correlation of  intra-feature and inter-feature, achieving the optimization of primary feature based on itself and associated features.
Take as an example TripleIT $T_2$, which is shown in Fig. \ref{fig:TripleIT}.

At first, all the features with different resolutions  are divided into many tokens, and each token is 1$\times$1$\times$128 size.
Second, an intra-feature self-attention (SA) layer is applied on primary feature. %Here, Query ($Q$), Key ($K$), Value ($V$) come from main feature itself.
It will comprehensive explore intra-feature relation with global receptive field.
Third, an inter-feature cross-attention layer is used between self-boosting primary feature and its associate features.
%Here, $Q$ comes from self-optimized main feature. $K$ and  $V$ come from associate features.
It will further excavate inter-feature relation and obtain long-range dependency across multiple resolutions.
Finally, a feed-forward (FF) layer is applied.
The process can be described as:
\begin{equation}
T_i=\textit{FF}\left(\textit{CA}\left(\textit{SA}\left(f^o_i\right),f^{asso}_i\right)\right)
\end{equation}
where $\textit{SA}\left(\cdot\right)$ is intra-feature self-attention, $\textit{CA}\left(\cdot\right)$ is inter-feature cross-attention, $f^{asso}_i$ is associated features of $f^o_i$ and $\textit{FF}\left(\cdot\right)$ is feedforward network (FFN).  Each operation follows by a residual connection and normalization layer.

The intra-feature self-attention uses the primary feature as input $Q$, $K$, and $V$, and output self-boosting primary feature.
The inter-feature cross-attention adopts self-boosting primary feature as $Q$, meanwhile uses the associated features as $K$ and $V$.
The intra-feature self-attention and inter-feature cross-attention are both implemented by efficient attention\cite{shen2021efficient} for less memory and computational costs.
In the inter-feature cross-attention layer, since the length of patches of primary feature and associated features are different, position embedding\cite{carion2020end} is used to avoid the position errors between tokens for learning the attention weight.%helps to consider the spatial information between any two positions in $Q$ and $K$ for learning the attention weight.

By stacking $L=2$ above architectures, the primary feature is refined based on its intra-feature self-relationship, and further optimized by useful information of all the associated features.
\section{Experiments}
\subsection{Datasets}\label{sec:Datasets}
\textbf{RGB-D SOD dataset} includes NLPR~\cite{peng2014rgbd} with 1,000 images, NJU2K~\cite{ju2014depth} with 1,985 images, STERE~\cite{niu2012leveraging} with 1,000 images, DES~\cite{cheng2014depth} with 135 images, SIP~\cite{fan2020rethinking} with 929 images, and DUT~\cite{piao2019depth} with 1,200 images.
\textbf{RGB-T SOD dataset} includes VT821\cite{wang2018rgb}, VT1000\cite{tu2019rgb}, and VT5000\cite{tu2020rgbt}.
\textbf{LF SOD dataset} includes LFSD\cite{li2014saliency} with 100 light field data, HFUT-Lytro\cite{zhang2017saliency} with 255 samples, DUTLF-FS\cite{wang2019deep} with 1,462 light field images.
%Each light field consists of an all-in-focus image, several focal slices focused at different depths and a pixel-wise ground truth of all-in-focus image.
%Moreover, we ensure the number of focal slices in each scene to be 12 by inserting the arbitrarily copied focal slices for coding requirements while preserving the original order.
\subsection{Evaluation metrics}
\textbf{Precision-recall (PR) curve~\cite{borji2015salient}} reflects the relation of precision and recall. The best model has both the best accuracy and the best recall.
\textbf{S-measure~\cite{fan2017structure}} evaluates structural similarity and emphasizes structural integrity.
\textbf{F-measure~\cite{achanta2009frequency}} is the adaptive value presentation of the PR curve.
\textbf{E-measure~\cite{fan2018enhanced}} captures adaptive global and local similarity.
\textbf{Mean absolute error (MAE)~\cite{perazzi2012saliency}} measures per-pixel absolute difference.

\subsection{Implementation details}
We use an NVIDIA RTX 3090 GPU to train our model.
The training set of RGB-D SOD comprises 2,185 paired RGB and depth images from NJU2K and NLPR. When testing the DUT dataset, our training set adds the DUT training set with 800 image pairs. The training set of RGB-T SOD includes 2,500 image pairs in VT5000. The training set of LF SOD consists of 100 samples in HFUT-Lytro and 1,000 samples in DUTLF-FS. The rest samples involved in Section \ref{sec:Datasets} are tested.
HRFormer in the primary stream and ResNet18 in the auxiliary stream use pre-trained  parameters. The rest of parameters are initialized to PyTorch default settings.
The loss function adopts pixel position aware loss\cite{wei2020f3net}.
During training, data enhancement, i.e., random flipping, rotating and border clipping, are conducted. The size of input image is set as 224$\times$224. The batch size is 15 and the initial learning rate is 5e-5.  The model converges within 300 epochs.
\subsection{Comparisons with SOTAs}
\subsubsection{RGB-D SOD}
There are many RGB-D SOD methods which  surpass previous methods in the last year. Therefore, we compare
our model with these pioneering works published in 2021, including JL-DCF\cite{fu2021siamese},
%MobileSal\cite{wu2021mobilesal},
CDNet\cite{jin2021cdnet},
DPANet\cite{chen2020dpanet},
HAINet\cite{li2021hierarchical},
DSNet\cite{wen2021dynamic},
CCAFNet\cite{zhou2021ccafnet},
EBFSP\cite{huang2021employing},
MMNet\cite{gao2021unified},
RD3D\cite{chen2021rd3d},
TriTransNet\cite{liu2021tritransnet},
DCF\cite{ji2021calibrated},
DSA2F\cite{sun2021deep},
VST\cite{liu2021visual},
SPNet\cite{zhou2021specificity},
EBMG\cite{zhang2021learning}, and SwinNet\cite{liu2021swinnet}.

\begin{figure*}[!htp]
\centering
\begin{tabular}{ccc}
\includegraphics[width = 0.33\textwidth]{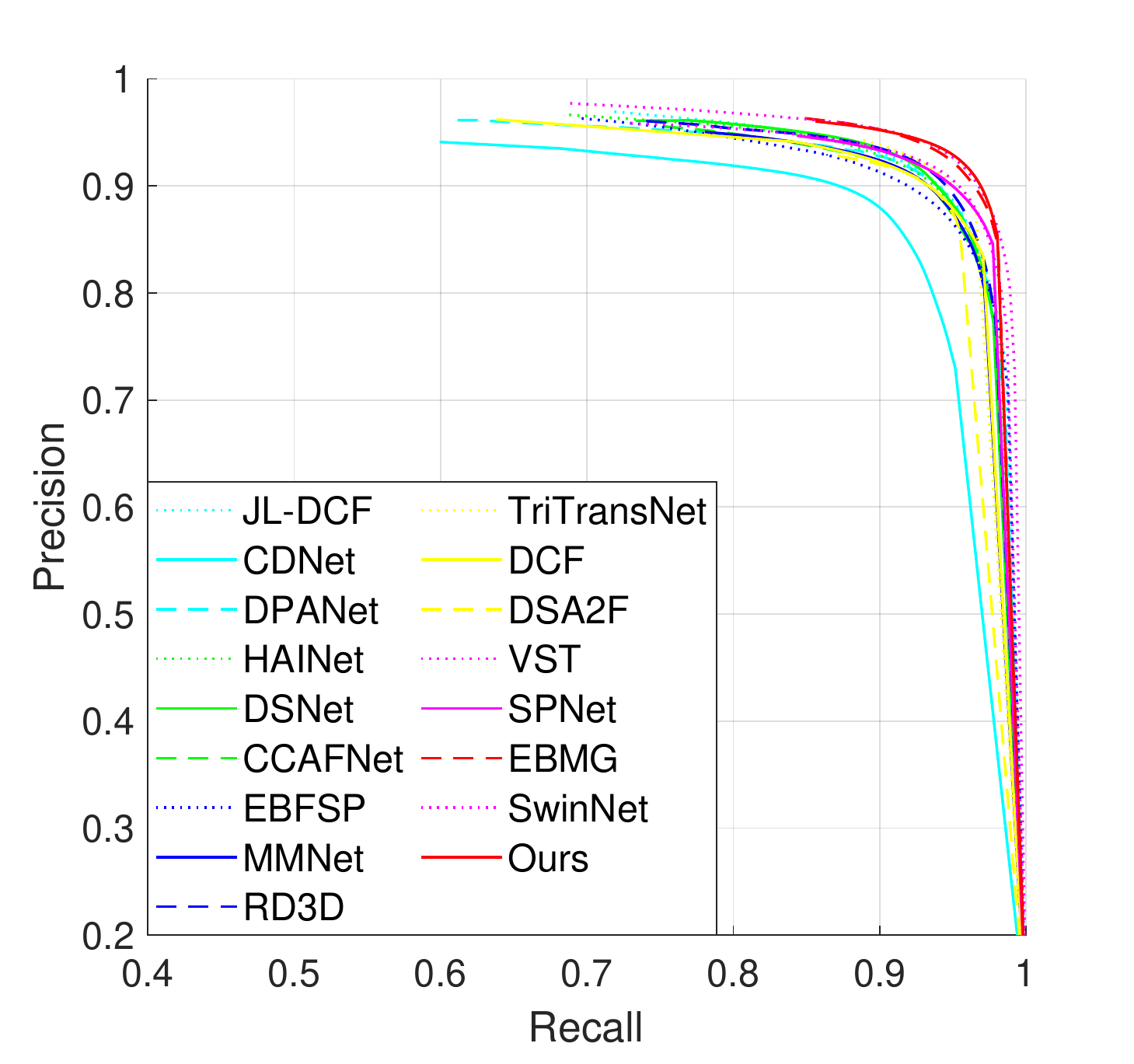}&\includegraphics[width = 0.33\textwidth]{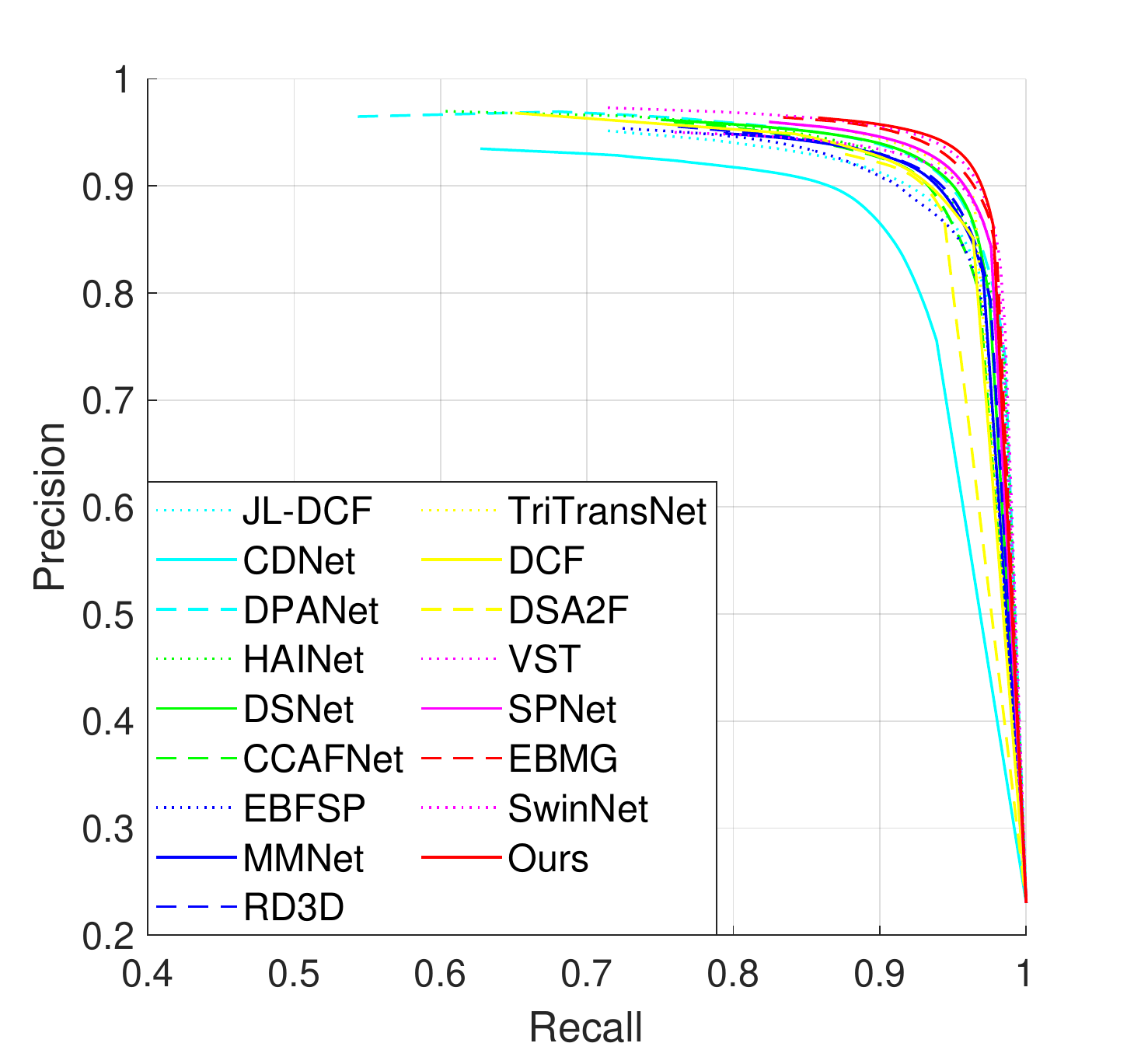}&\includegraphics[width = 0.33\textwidth]{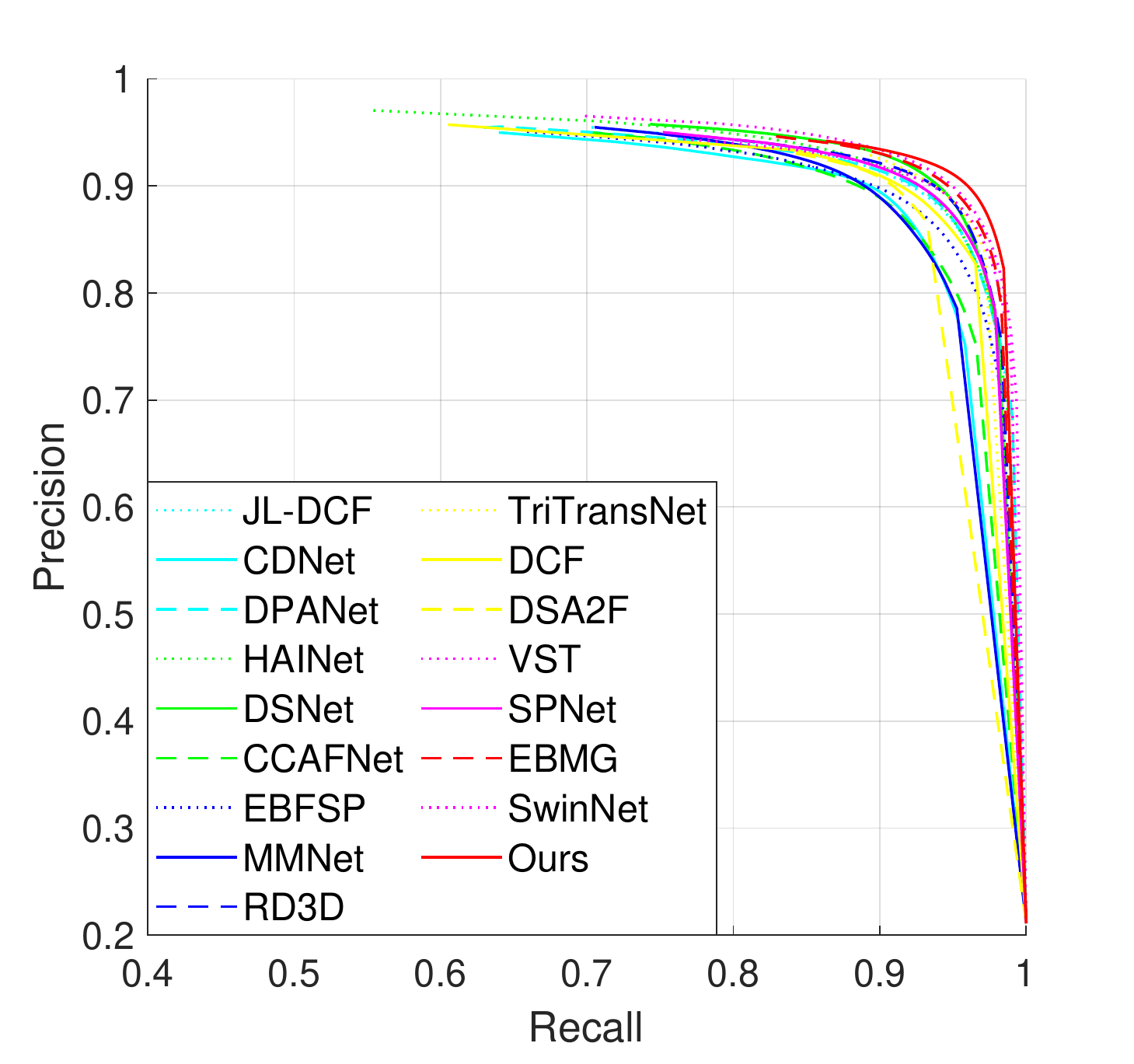}\\
(a) NLPR dataset&(b) NJU2K dataset&(c) STERE dataset\\
\end{tabular}
\begin{tabular}{ccc}
\includegraphics[width =
0.33\textwidth]{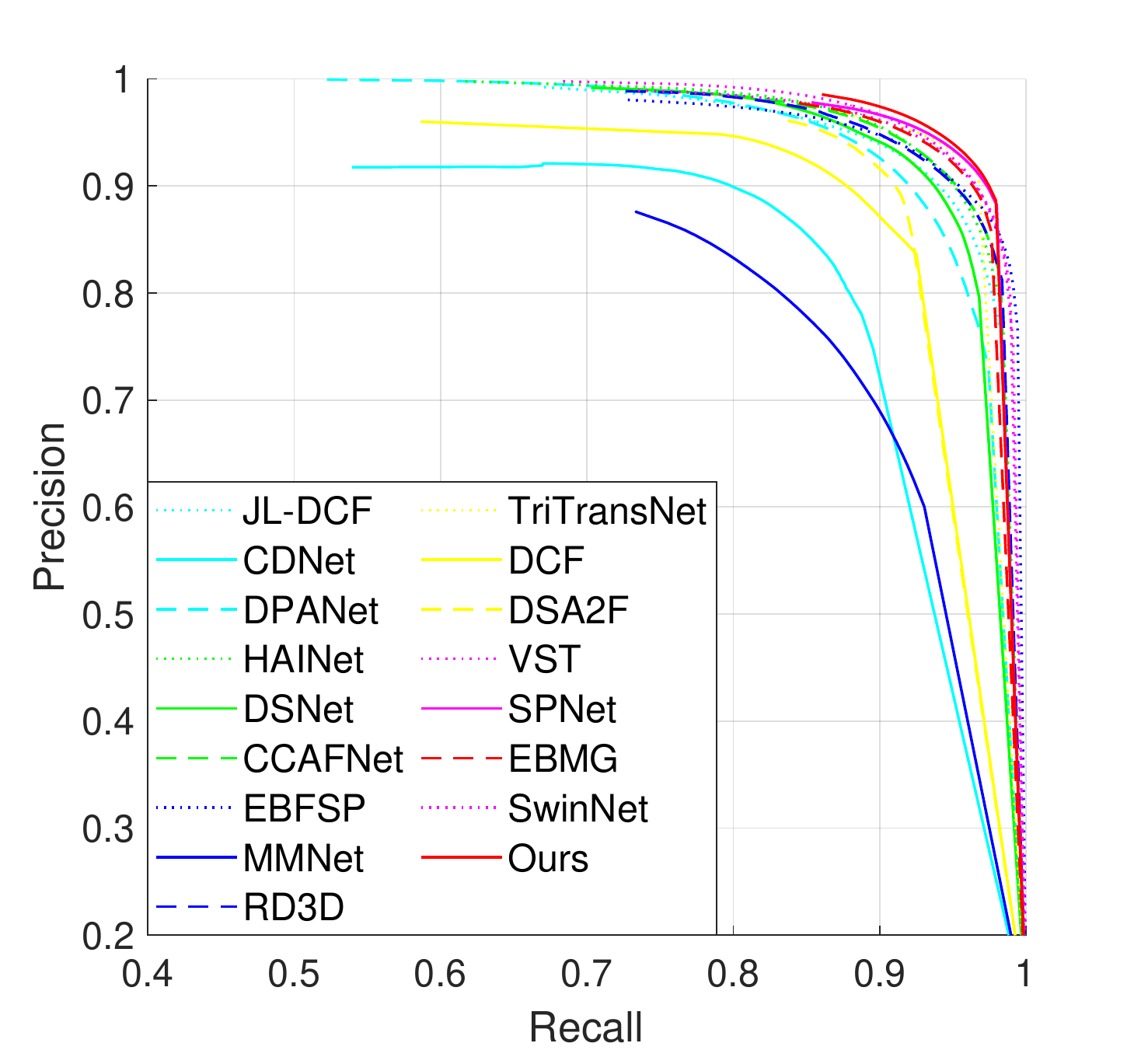}&\includegraphics[width = 0.33\textwidth]{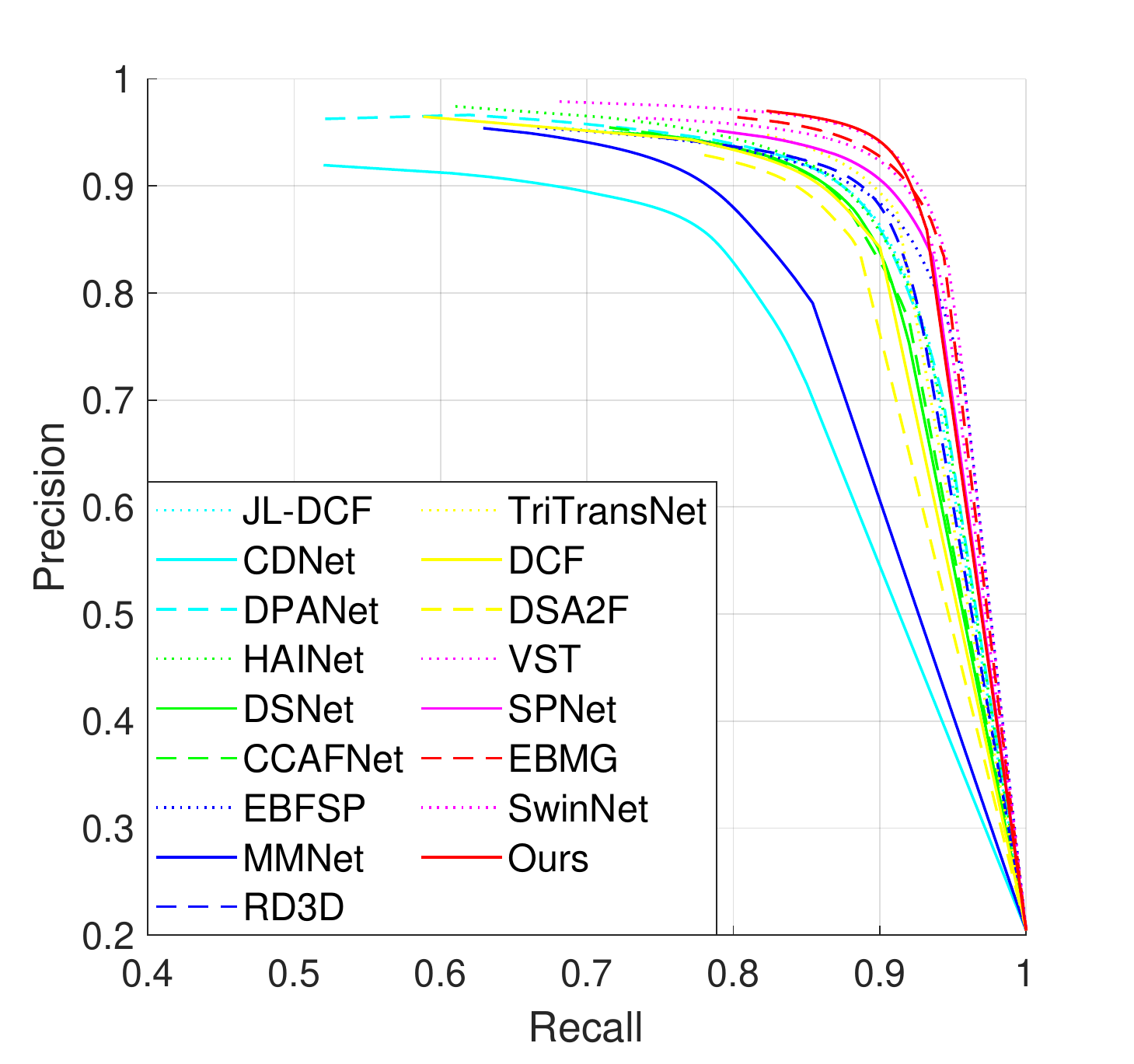}&\includegraphics[width = 0.33\textwidth]{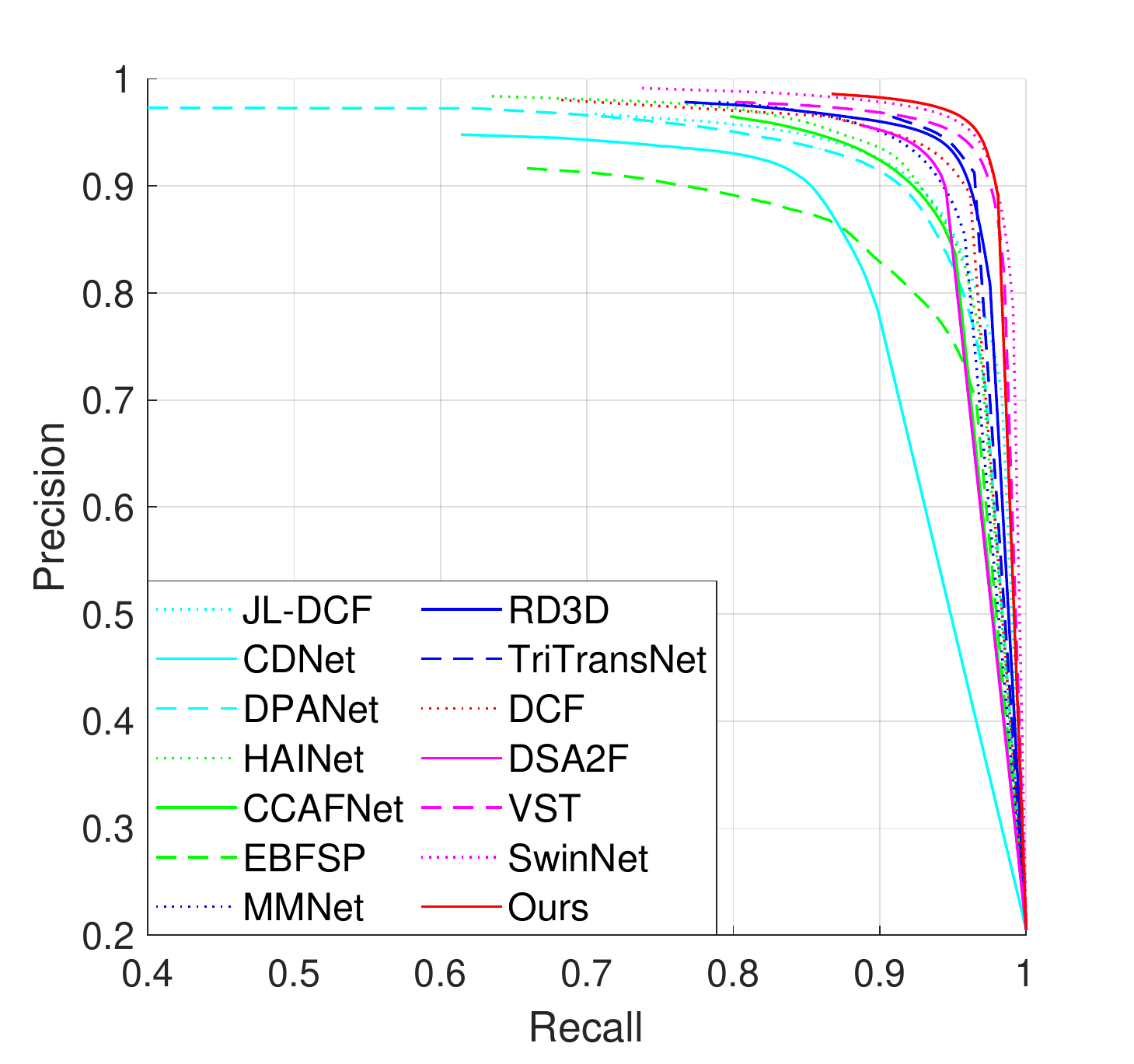}\\
(d) DES dataset &(e) SIP dataset&(f) DUT dataset\\
\end{tabular}
\caption{The comparison of P-R curves  on six RGB-D datasets.}
\label{fig:RGBDPRComparison}
\end{figure*}

\textbf{Quantitative Evaluation.} The PR curve in Fig. \ref{fig:RGBDPRComparison} shows accuracy and recall of different models.
Our curves are better than the others on all datasets. It benefits from the initial selective fusion of two modalities  and the integration of  multi-resolution output features of HRFormer.
Four evaluation metrics in Table. \ref{tab:RGBDcomparison}
also give the same results when compared with pioneering works published in CVPR, ICCV, NeurIPS, and so on. Most evaluation metrics are superior to the others. It indicates that our method is better than the others, and comparable with EBMG\cite{zhang2021learning} and SwinNet\cite{liu2021swinnet}. Moreover, from the bottom of the table, we can find our method is superior in the computation cost. It has the fewer parameters and lower FLOPs than most compared methods.

    \begin{table*}[!htp]
  \centering
  \fontsize{6}{10}\selectfont
  \renewcommand{\arraystretch}{0.4}
  \renewcommand{\tabcolsep}{0.4mm}
  \scriptsize
  %\captionsetup{labelformat=empty}
  \caption{The comparison of S-measure, adaptive F-measure, adaptive E-measure, MAE of different RGB-D SOD models.  ``-" means that the method does not release test results for this dataset or code. }
\label{tab:RGBDcomparison}
  \scalebox{0.9}{
  \begin{tabular}{c|c|cccccccccccccccc|c}
  \hline\toprule
    Datasets& Metric &
       JL-DCF    & CDNet&DPANet & HAINet &DSNet &CCAFNet &EBFSP &MMNet &RD3D  &TriTransNet &DCF &DSA2F &VST&SPNet&EBMG&SwinNet&\multirow{2}{*}{HRTransNet}  \\
       & & TPAMI21    & TIP21& TIP21 & TIP21 &TIP21 &TMM21 &TMM21 &TCSVT21 &AAAI21  &ACM MM21 &CVPR21 &CVPR21 &ICCV21&ICCV21&NeurIPS21&TCSVT21& \\
       &&\cite{fu2021siamese}& \cite{jin2021cdnet}&\cite{chen2020dpanet}& \cite{li2021hierarchical}& \cite{wen2021dynamic}& \cite{zhou2021ccafnet}& \cite{huang2021employing}& \cite{gao2021unified}& \cite{chen2021rd3d}& \cite{liu2021tritransnet}& \cite{ji2021calibrated}& \cite{sun2021deep}& \cite{liu2021visual}& \cite{zhou2021specificity}& \cite{zhang2021learning}&\cite{liu2021swinnet}& Ours\\

  \midrule

  \multirow{3}{*}{\textit{NLPR}}
    & $S\uparrow$    & .925  & .902  &.928& .924  & .926 & .922 &.915 &.925 &.933  &.928 &.923 &.918 &.931 &.927&.938&.941&$\textbf{.942}$\\

    & $F_\beta \uparrow$      & .878  & .848 &.875 & .897  & .886 & .880  &.897&.889&.886&.909 &.876& .892 &.886 &.904 &\textbf{.920}&.908&.919     \\

    & $E_{\xi}\uparrow$       & .954& .935&.953& .957& .955&.951&.952&.950&.958 &.960 & .950 &.950 &.954 &.958&.967&.967&$\textbf{.969}$\\

    & MAE$\downarrow$         & .022 & .032 & .024& .024  & .024 & .026 & .026&.024&.022 &.020 &.024& .024 &.023 &.021&.018&.018&\textbf{.016}\\
    \midrule

  \multirow{3}{*}{\textit{NJU2K}}
    & $S\uparrow$       & .903  & .885&.921 & .912 & .921  & .910  & .903 & .911     &.928 &.920 &.912&.904 &.922 &.924&.929&.935&$\textbf{.933}$ \\

    &$F_\beta$$\uparrow$      & .884 & .866 &.887& .900 & .907  & .897  & .894 & .900  &.910  &.919 &.902& .898 &.899 &.917&.923&.922&$\textbf{.928}$ \\

    & $E_{\xi}\uparrow$        & .913  & .908 &.915& .922 & .920 & .920 & .907 & .919  & .920   &.925 & .924    &.922 &.914 &.932&\textbf{.935} &.934& .931  \\

    & MAE$\downarrow$            & .041 & .048 &.035 & .038  & .034 & .037 &.039&.038&.033 &.030 & .035& .039 &.034 &.028&.028&.027&$\textbf{.026}$ \\

    \midrule

  \multirow{3}{*}{\textit{STERE}}
    & $S\uparrow$     & .903  & .896 &.909& .907 &.915  & .892   & .900 & .891 &.914  &.908 &.902 &.897 &.913 &.907&.916&.919&$\textbf{.921}$\\

    & $F_\beta$$\uparrow$      & .869  & .873&.875 & .885 & .894 & .869   & .870 & .880  &.881  &.893 &.884 &.893 &.878 &.888&.899&.893&\textbf{.904}\\

    & $E_{\xi}\uparrow$       & .918  & .922&.920 & .925 & .929 & .921 & .912 & .924  & .921  &.927 &.929 &.927 &.917 &$\textbf{.930}$&$\textbf{.930}$&.929&$\textbf{.930}$ \\

    & MAE$\downarrow$            & .040  & .042&.040 & .040 & .036  & .044   & .045 & .045 &.039  &.033 &.039 &.039 &.038 &.037&.032&.033&$\textbf{.030}$\\
    \midrule

  \multirow{3}{*}{\textit{DES}}
    & $S\uparrow$    & .931  & .876 &.919 & .935  & .927 & .938 &.937 &.830 &.950  &.943 &.904 &.916 &.943 &.944&.938&.945&$\textbf{.947}$\\

    & $F_\beta$$\uparrow$      & .900  & .840 &.900& .924 & .910  & .915   & .913 & .746&.915 &.936 &.876&.901 &.917 &.935&.928&.926&\textbf{.938}\\

    & $E_{\xi}\uparrow$        & .969 & .921 &.963& .974 & .970 & .975 & .974 & .893 & .979    & .981 &.950 & .955 &.979 &.983 &.977&.980&$\textbf{.983}$ \\

    & MAE$\downarrow$             & .020 & .034 &.023& .018 & .021  & .018   & .018 & .058  &.017 &\textbf{.014} &.024&.023 &.017 &\textbf{.014}&.016&.016&\textbf{.014}\\
    \midrule

  \multirow{3}{*}{\textit{SIP}}
    & $S\uparrow$    & .880  & .823 &.883 & .880  & .876 & .877 &.885 &.836 &.892  &.886 &.875 &.862 &.904 &.894&.902&\textbf{.911}&.909\\

    & $F_\beta$$\uparrow$      & .873  & .805 &.865& .875 & .865  & .864   & .869 & .839&.883 &.892 &.875&.865 &.895 &.893&.908&.912&\textbf{.916}\\

    & $E_{\xi}\uparrow$        & .921  & .880&.921 & .919 & .910 & .915 & .917 & .882 & .924    &.924 & .920 &.908 &.937 &.930&.938&\textbf{.943}&\textbf{.943} \\

    & MAE$\downarrow$             & .049  & .076 &.051& .053 & .052  & .054   & .049 & .075  &.046 &.043 &.052&.057 &.040 &.043&.037&\textbf{.035}&\textbf{.035}\\
    \midrule

  \multirow{3}{*}{\textit{DUT}}
    & $S\uparrow$    & .906  & .880 &.899 & .909  & - & .904 &.858 &.920 &.936  &.934 &.924 &.921 &.943 &-&-&.949&\textbf{.951}\\

    & $F_\beta$$\uparrow$      & .882  & .874&.881 & .905 & -  & .903   & .841 & .919&.925 &.936&.925 &.926 &.931&-&-&.944&\textbf{.955}\\

    & $E_{\xi}\uparrow$        & .931  & .918 &.923& .937 &- & .940 & .890 & .951 & .953    &.957 &.952 &.950 &.960&-&-&.968&\textbf{.972} \\

    & MAE$\downarrow$             & .043 & .048 & .048 & .038 &-  & .037   & .067 & .032  &.030 &.025 &.030&.030 &.025 &-&-&.020&\textbf{.018}\\
\midrule

\multicolumn{2}{c|}{\centering Params(M)$\downarrow$}

	&143.5	&\textbf{32.9}&	92.4&	59.8&-&-&-&		64.1	&46.9	&138.7&	108.5	&36.5	&83.0	&67.9&	88.8	&198.7&	58.9\\
\midrule
\multicolumn{2}{c|}{\centering FLOPs(G)$\downarrow$}
 &	861.2&	72.1&	58.9&	181.4&-&-&-&				42.4	&26.8&	292.3&	54.0	&364.4	&31.0	&150.3&	303.9	&124.3	&\textbf{17.1}\\

%\multicolumn{2}{c|}{\centering Params(M)}
%	&143.5	&32.9&	92.4&	59.8&-&-&-&		64.1	&46.9	&138.7&	108.49	&36.5	&83.0	&67.9&	88.8	&198.7&	58.9\\
%\midrule
%\multicolumn{2}{c|}{\centering FLOPs(G)}&	861.2&	72.1&	58.9&	181.4&-&-&-&				42.4	&26.8&	292.3&	54.0	&364.4	&31.0	&150.3&	303.9	&124.3	&17.1\\

  \bottomrule
  \hline
  \end{tabular}}
\end{table*}

\begin{figure*}[!htp]
	\centering \includegraphics[width=1\textwidth]{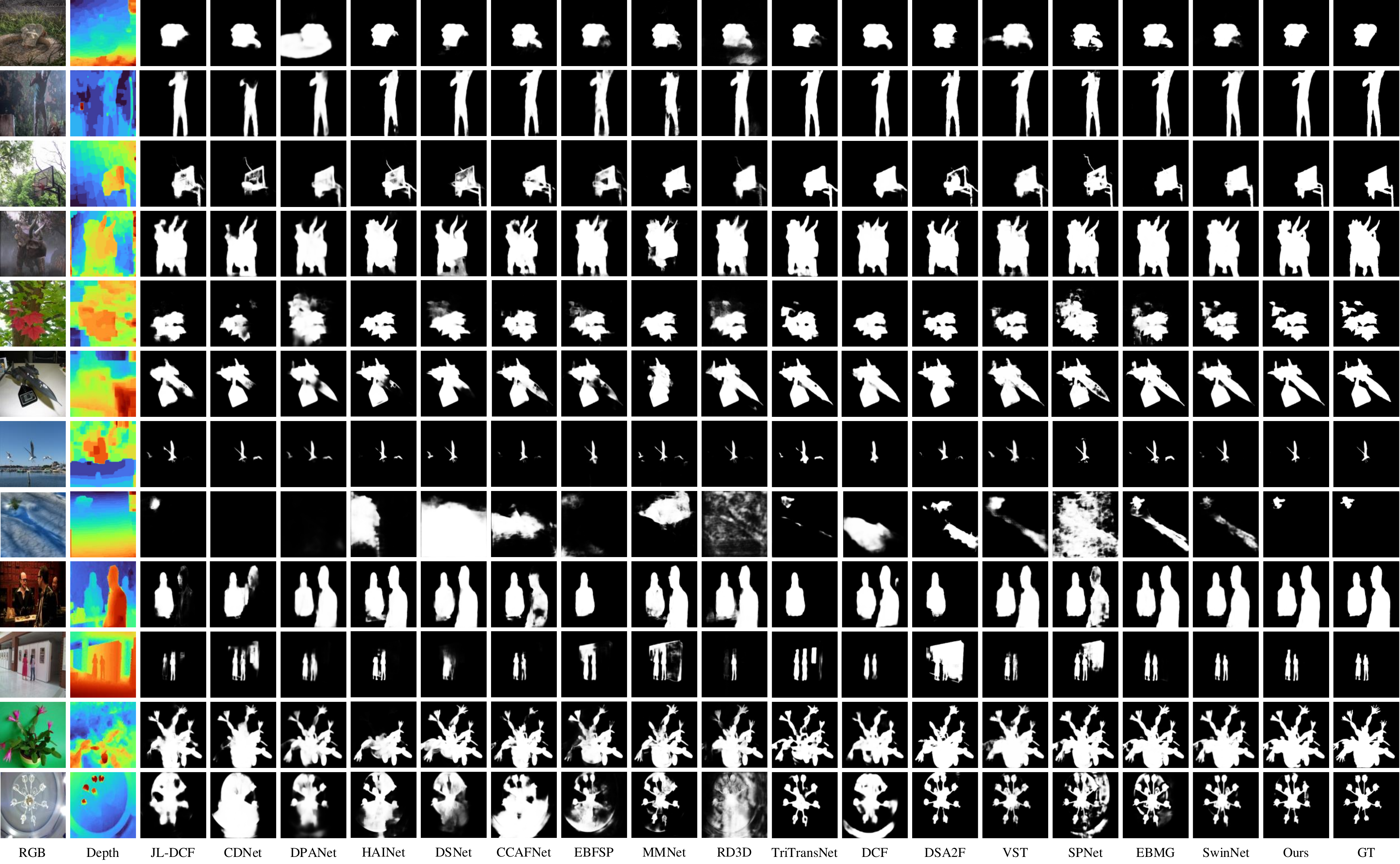}
	\caption{Visual comparison of RGB-D SOD. Our HRTransNet is outstanding in some scenes: indistinguishable foreground object ($1^{st}_{}$-$2^{nd}_{}$ rows), complex scene ($3^{rd}_{}$-$4^{th}_{}$ rows),  low-quality depth image ($5^{th}_{}$-$6^{th}_{}$ rows), small objects ($7^{th}_{}$-$8^{th}_{}$ rows), multiple objects ($9^{th}_{}$-$10^{th}_{}$ rows), and fine-grained objects ($11^{th}_{}$-$12^{th}_{}$ rows).}
\label{fig:RGBDVisualCompare}
\end{figure*}

\textbf{Qualitative Evaluation.} The visual examples in Fig. \ref{fig:RGBDVisualCompare} show the performance of our model in some scenes: indistinguishable foreground object ($1^{st}_{}$-$2^{nd}_{}$ rows), complex scene ($3^{rd}_{}$-$4^{th}_{}$ rows),  low-quality depth image ($5^{th}_{}$-$6^{th}_{}$ rows), small objects ($7^{th}_{}$-$8^{th}_{}$ rows), multiple objects ($9^{th}_{}$-$10^{th}_{}$ rows), and fine-grained objects ($11^{th}_{}$-$12^{th}_{}$ rows).
It is due to the proposed several modules.
For example, supplementary modality injection module reduces the influence of low-quality depth images by effectively filtering out depth noise. It ensures good performance  in the scene with low-quality depth image.
For example, the dual-direction short connection fusion module combines the merits of multi-resolution features, showing  superior performance in detecting fine-grained details.
For example, HRFormer with transformer blocks maintains larger receptive fields than the original HRNet. It ensures powerful detection ability in small objects and multiple objects, complex scenes, etc.

\subsubsection{RGB-T SOD}
Compared with RGB-D SOD,  RGB-T SOD is developed slower.
Comparison methods include
MTMR\cite{wang2018rgb}, %(VT821)(Wang)RGB-T saliency detection benchmark: Dataset, baselines, analysis and a novel approach
M3S-NIR\cite{tu2019m3s}, %(Tu)M3S-NIR: Multi-modal multi-scale noise-insensitive ranking for RGB-T saliency detection \\
SGDL\cite{tu2019rgb}, %(VT1000)(Tu)RGB-T image saliency detection via collaborative graph learning
ADF\cite{tu2020rgbt},%(VT5000)(Tu)Rgbt salient object detection: a Consistent Feature Fusion Network for RGB-T Salient Object Detection
MIDD\cite{tu2021multiinteractive},
ECFFNet\cite{zhou2021ecffnet}, %(Zhou)ECFFNet: Effective and large-scale dataset and benchmark\\
MMNet\cite{gao2021unified},
CSRNet\cite{huo2021efficient},
CGFNet\cite{wang2021cgfnet},
and SwinNet\cite{liu2021swinnet}, which are published in recent five years.

\begin{figure*}[!htp]
\centering
\begin{tabular}{ccc}
\includegraphics[width = 0.33\textwidth]{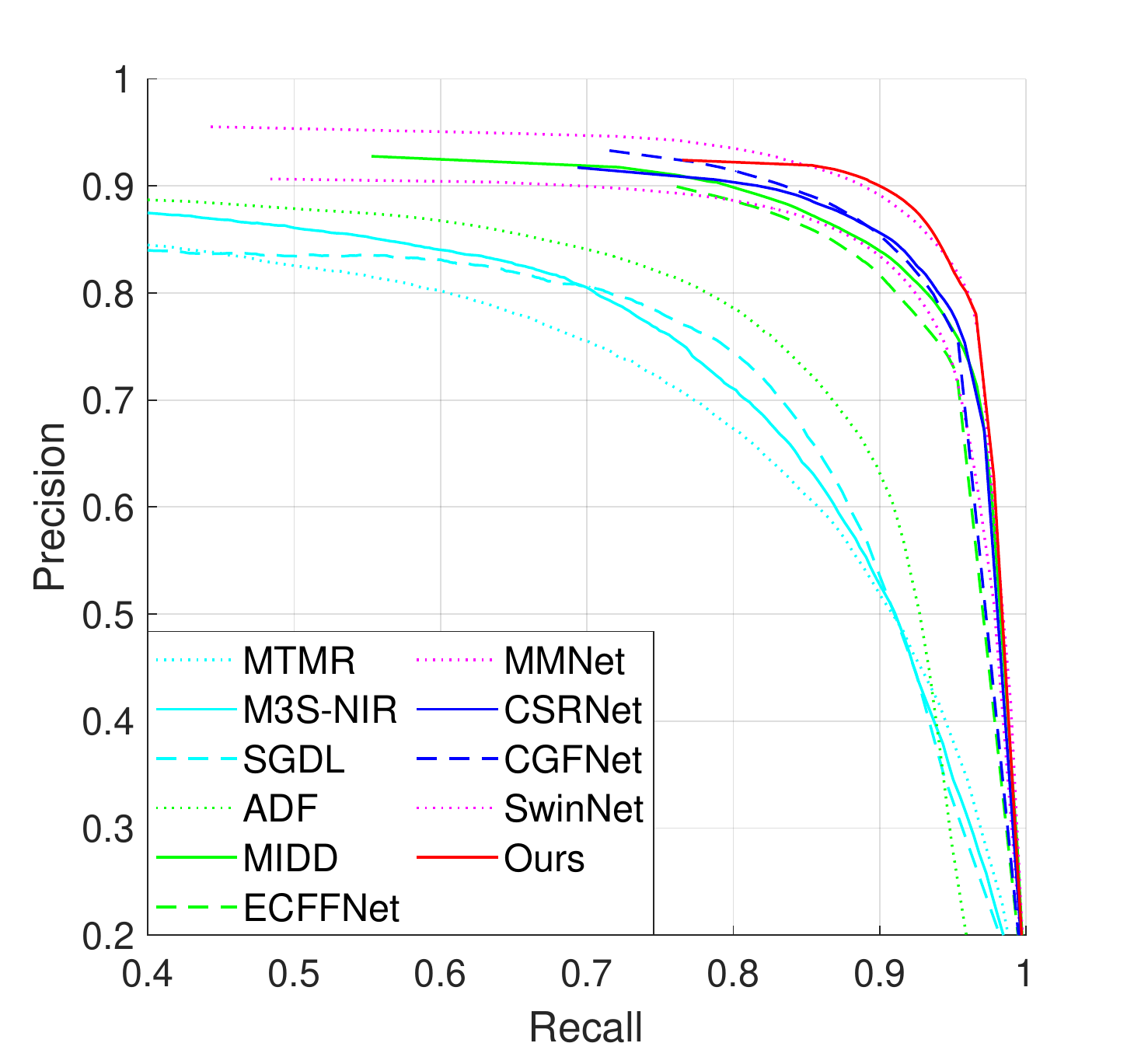}&\includegraphics[width = 0.33\textwidth]{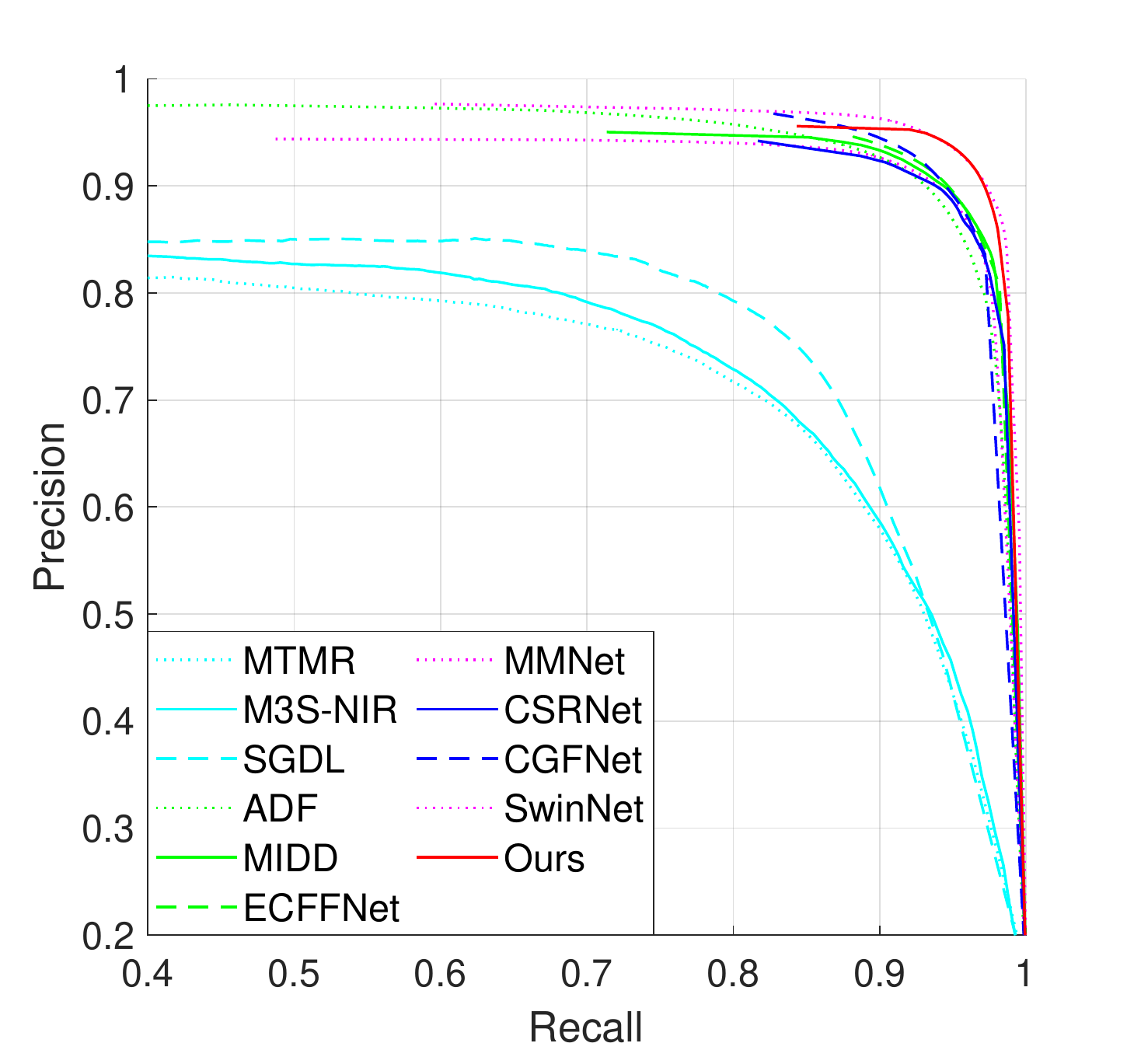}&\includegraphics[width = 0.33\textwidth]{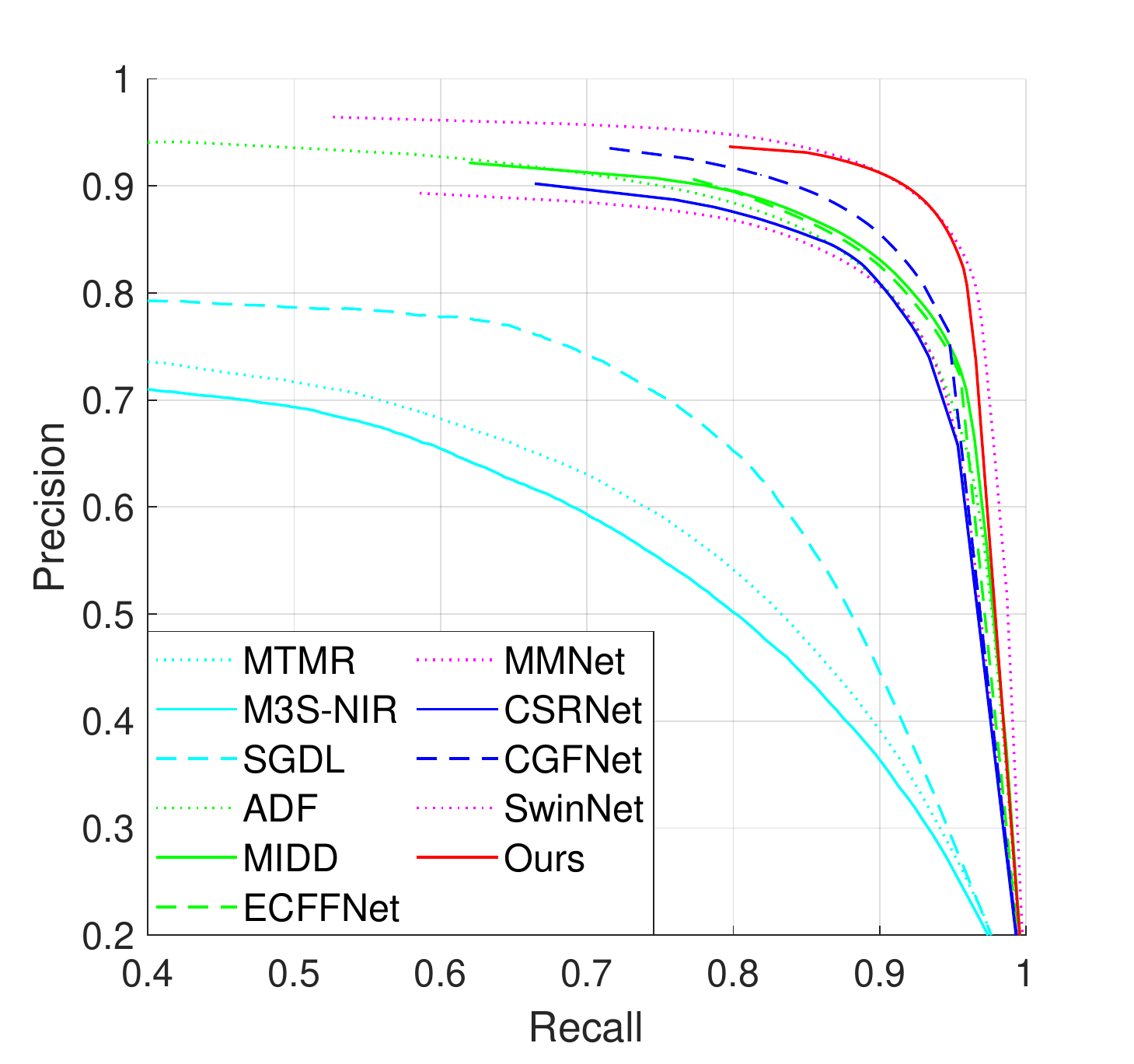}\\
(a) VT821 dataset&(b) VT1000 dataset&(c) VT5000 dataset\\
\end{tabular}
\caption{The comparison of P-R curves  on three RGB-T datasets.}
\label{fig:RGBTPRComparison}
\end{figure*}

\textbf{Quantitative Evaluation.} The PR curve in Fig. \ref{fig:RGBTPRComparison} shows that our method wins the others with a great margin. Four evaluation metrics in Table. \ref{tab:RGBTcomparison} also reveal the  outstanding performance of our models.  %The MAE value is improved about 0.1 on average.
The figure and table both indicate our method is robust when applied in RGB and thermal image pairs.

\begin{table*}[!htp]
  \centering
  \fontsize{6}{10}\selectfont
  \renewcommand{\arraystretch}{0.4}
  \renewcommand{\tabcolsep}{0.4mm}
  \scriptsize
  %\captionsetup{labelformat=empty}
  \caption{The comparison of S-measure, adaptive F-measure, adaptive E-measure, MAE of different RGB-T SOD models. }
\label{tab:RGBTcomparison}
  \scalebox{1}{
  \begin{tabular}{c|c|cccccccccc|c}
  \hline\toprule
    Datasets& Metric &
       MTMR\cite{wang2018rgb} &M3S-NIR\cite{tu2019m3s} &SGDL\cite{tu2019rgb} &ADF\cite{tu2020rgbt} &MIDD\cite{tu2021multiinteractive} &ECFFNet\cite{zhou2021ecffnet} &MMNet\cite{gao2021unified} &CSRNet\cite{huo2021efficient}&CGFNet\cite{wang2021cgfnet}&SwinNet\cite{liu2021swinnet}&HRTransNet \\
       & &IGTA18 &MIPR19 &TMM19 &TMM22 &TIP21 &TCSVT21 &TCSVT21&TCSVT21&TCSVT21&TCSVT21&Ours  \\

  \midrule

  \multirow{3}{*}{\textit{VT821}}
    & $S\uparrow$    & .725 & .723 & .765  & .810  &.871 &.877&.875&.885&.881&.904&$\textbf{.906}$\\

    & $F_\beta \uparrow$    &.662 &.734 &.730& .716 &.804 &.810 &.798 &.830&.845&.847&\textbf{.853}     \\

    & $E_{\xi}\uparrow$       &.815 & .859 & .847 &.842 &.895 &.902&.893&.908&.912&.926&$\textbf{.929}$\\

    & MAE$\downarrow$         &.108 &.140 &.085& .077 &.045 &.034&.040&.038&.038&.030&\textbf{.026}\\
    \midrule

  \multirow{3}{*}{\textit{VT1000}}
    & $S\uparrow$      &.706 &.726 &.787&.910 &.915 &.923&.917&.918&.923&\textbf{.938}&$\textbf{.938}$ \\

    &$F_\beta$$\uparrow$     &.715 &.717 &.764& .847 &.882 &.876&.863&.877&$\textbf{.906}$&.896&.900 \\

    & $E_{\xi}\uparrow$       & .836 &.827 & .856    &.921 &.933 &.930&.924&.925 &.944&\textbf{.947}&.945  \\

    & MAE$\downarrow$          & .119 &.145 & .090& .034 &.027 &.021&.027&.024&.023&.018&$\textbf{.017}$ \\

    \midrule

  \multirow{3}{*}{\textit{VT5000}}
    & $S\uparrow$     &.680 &.652 &.750 &.863 &.867 &.874&.864&.868&.883&\textbf{.912}&$\textbf{.912}$\\

    & $F_\beta$$\uparrow$    &.595 &.575 &.672 &.778 &.801 &.806&.785&.810 &.851&.865&\textbf{.871}\\

    & $E_{\xi}\uparrow$     & .795 & .780 & .824 & .891 & .897 & .906 &.890&.905 &.922&.942& $\textbf{.945}$\\

    & MAE$\downarrow$       &.114 &.168 &.089 &.048 &.043 &.038&.043&.042&.035&.026&$\textbf{.025}$\\

  \bottomrule
  \hline
  \end{tabular}}

\end{table*}

\begin{figure*}[!htp]
	\centering	\includegraphics[width=\textwidth]{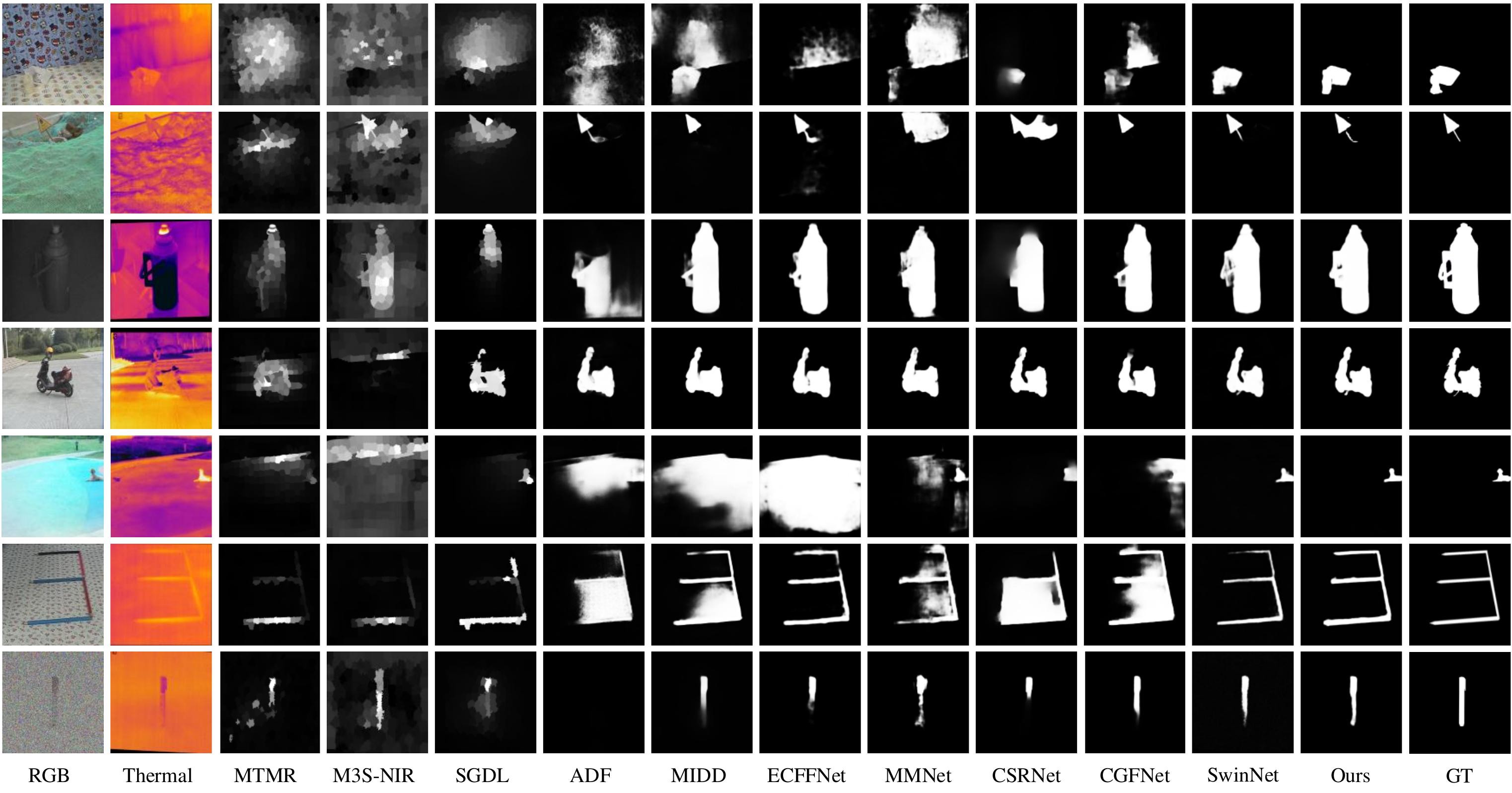}
	\caption{Visual comparison of RGB-T SOD. Our HRTransNet is outstanding in some scenes: indistinguishable foreground object ($1^{st}$ row), cluttered scene ($2^{nd}$ row), weak light condition ($3^{rd}$ row), low-quality thermal image ($4^{th}$ row), small objects ($5^{th}$ row), multiple objects ($6^{th}$ row), and scenes filled with noises ($7^{th}$ row).}
\label{fig:RGBTVisualCompare}
\end{figure*}

\textbf{Qualitative Evaluation.}
The visual examples in Fig. \ref{fig:RGBTVisualCompare} show the performance comparison of different models in some scenes: indistinguishable foreground object ($1^{st}$ row), cluttered scene ($2^{nd}$ row), weak light condition ($3^{rd}$ row), low-quality thermal image ($4^{th}$ row), small objects ($5^{th}$ row), multiple objects ($6^{th}$ row), and scenes filled with noises ($7^{th}$ row).
They suggest the effectiveness of proposed modules in the SOD task by the combination of color and thermal infrared information.

\subsubsection{LF SOD}
Compared with RGB-D and RGB-T SOD, LF SOD is studied by fewer researchers.
Comparison methods comprises seven methods published in recent four years, including MoLF\cite{zhang2019memory}, DLFS\cite{piao2019deep},  LFNet\cite{zhang2020lfnet}, ERNet\cite{piao2020exploit}, SA-Net\cite{zhang2021SANet}, DLGLRG\cite{liu2021light}, PANet\cite{piao2021panet}.
As we all know, LF image includes multi-focal, multi-view\cite{zhang2020multi}, and EPIs\cite{jing2021occlusion,fu2020light}.
We mainly discuss multi-focal LF whose input is an all-in-focus image and focal stack with not more than 12 focal slices.
%The focal stack consists of 12 focal slices focused on different depths.
When the number of focal slices is less than 12, we use an all-zero image to pad\cite{zhang2021SANet}.
Then all-in-focus image serves as the primary color feature, and multiple slices are concatenated to serve as the supplementary feature.

\begin{figure*}[!htp]
\centering
\begin{tabular}{ccc}
\includegraphics[width = 0.33\textwidth]{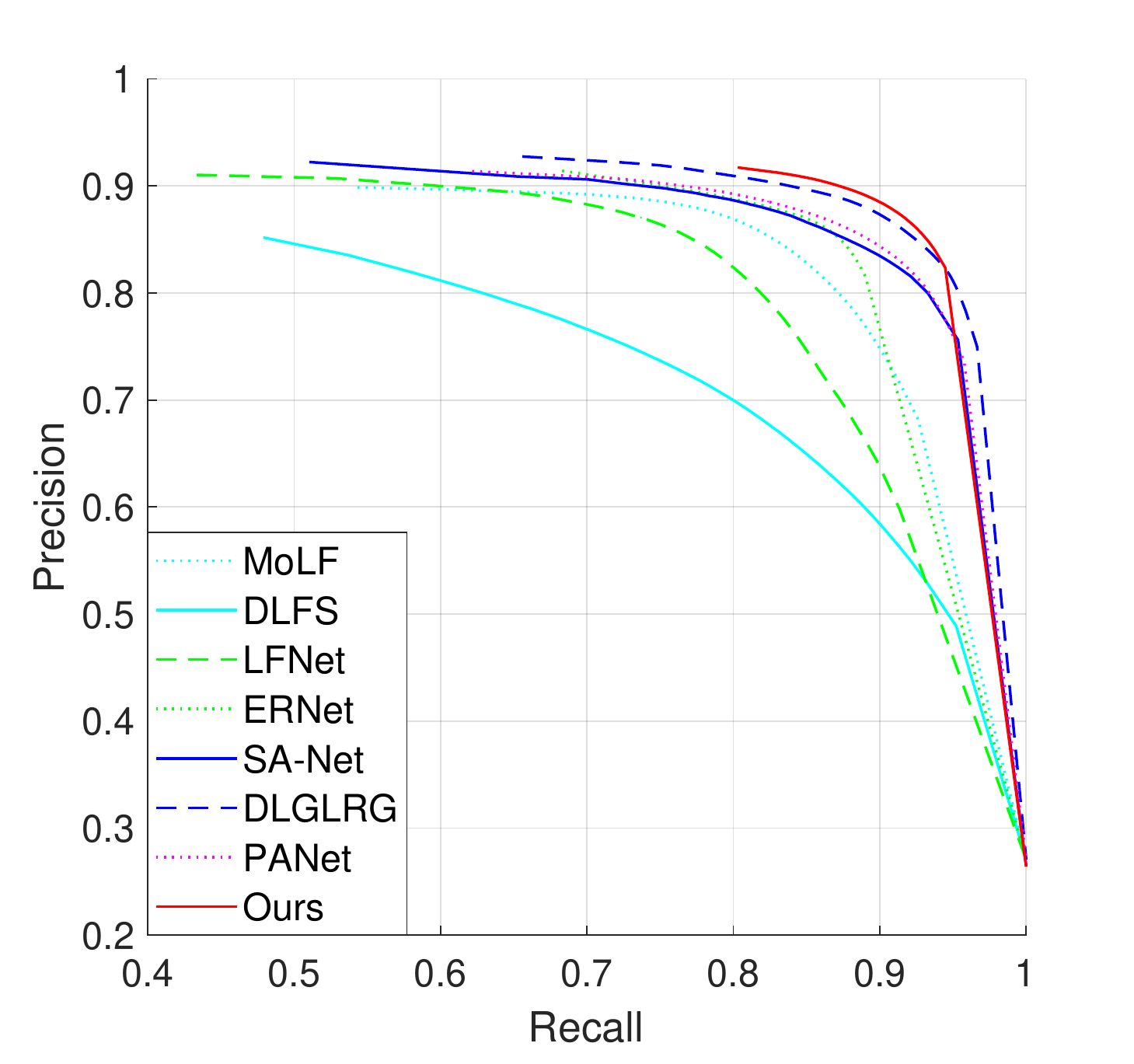}&\includegraphics[width = 0.33\textwidth]{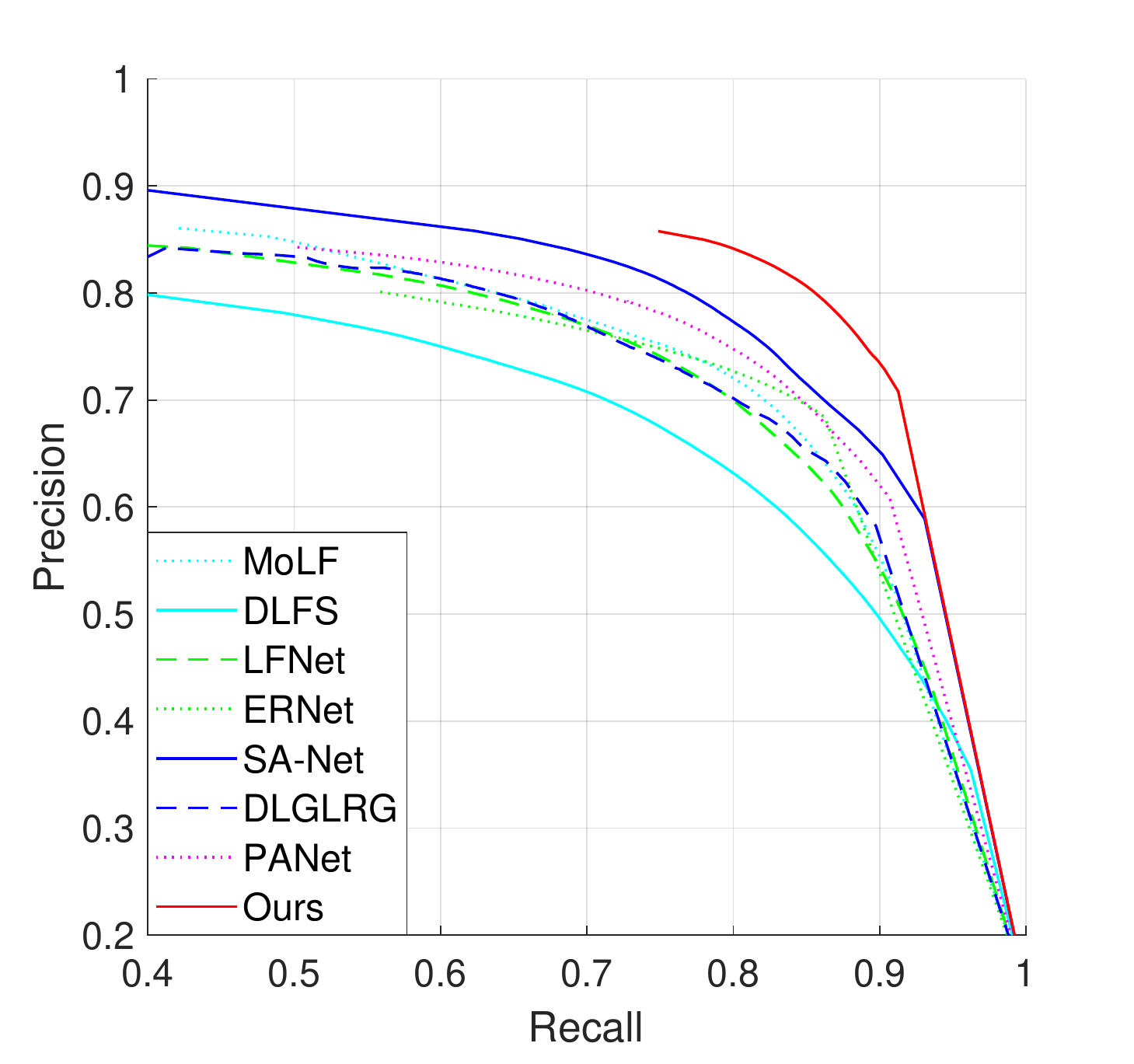}&\includegraphics[width = 0.33\textwidth]{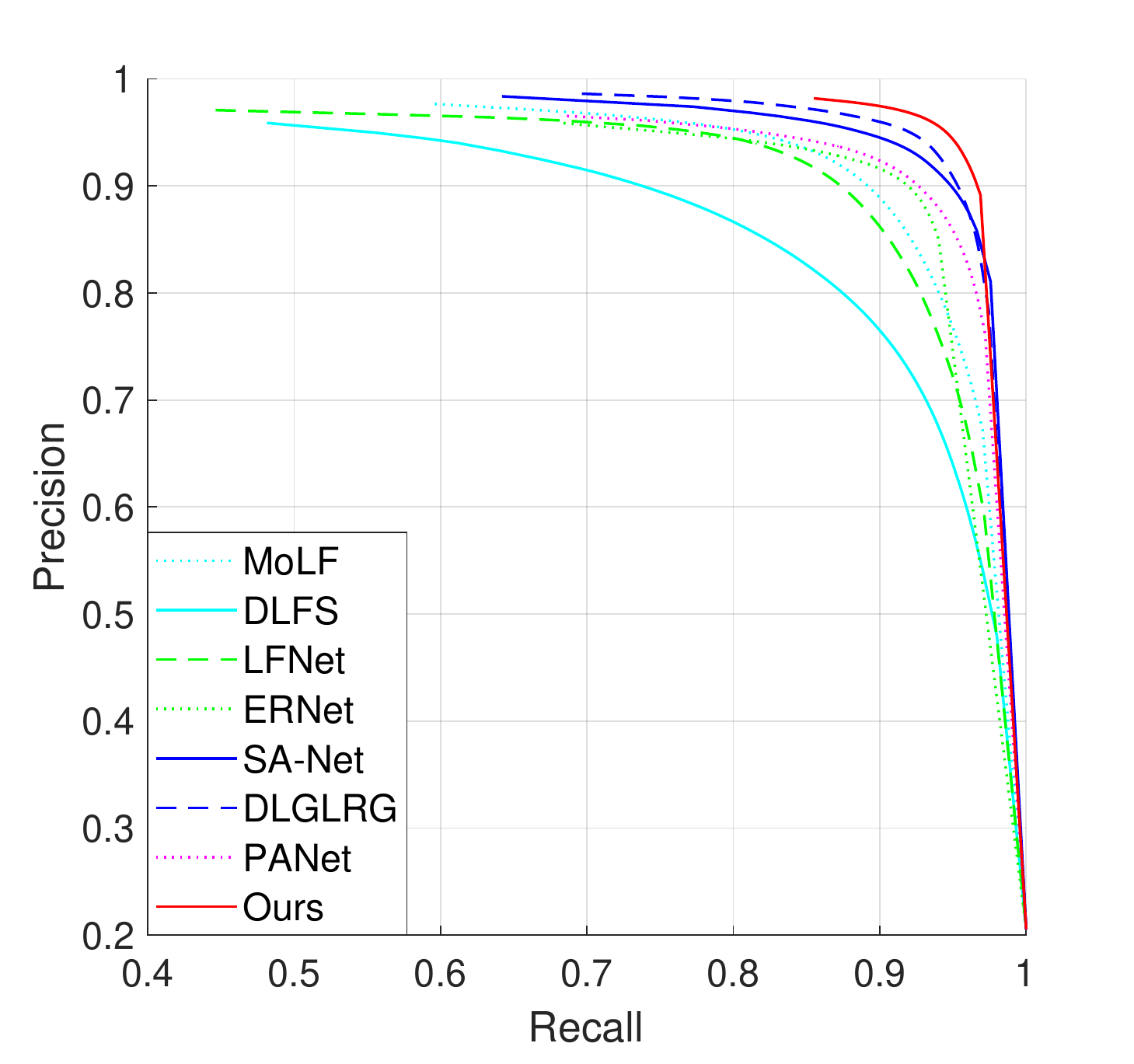}\\
(a) LFSD&(b) HFUT-Lytro&(c) DUTLF-FS\\
\end{tabular}
\caption{The comparison of P-R curves  on three light field datasets.}
\label{fig:LFPRComparison}
\end{figure*}

\textbf{Quantitative Evaluation.} The PR curve in Fig. \ref{fig:LFPRComparison} shows that our model outperforms the others in HFUT-Lytro and DUTLF-FS datasets, and is slightly  better than the others in LFSD dataset.
Four evaluation metrics in Table. \ref{tab:LFcomparison} shows a performance improvement.
The excellent performance is achieved only by concatenating focal slices. It also verifies the strong  compatibility of our model in processing the two-modality SOD tasks.

\begin{figure*}[!htp]
	\centering	\includegraphics[width=0.7\textwidth]{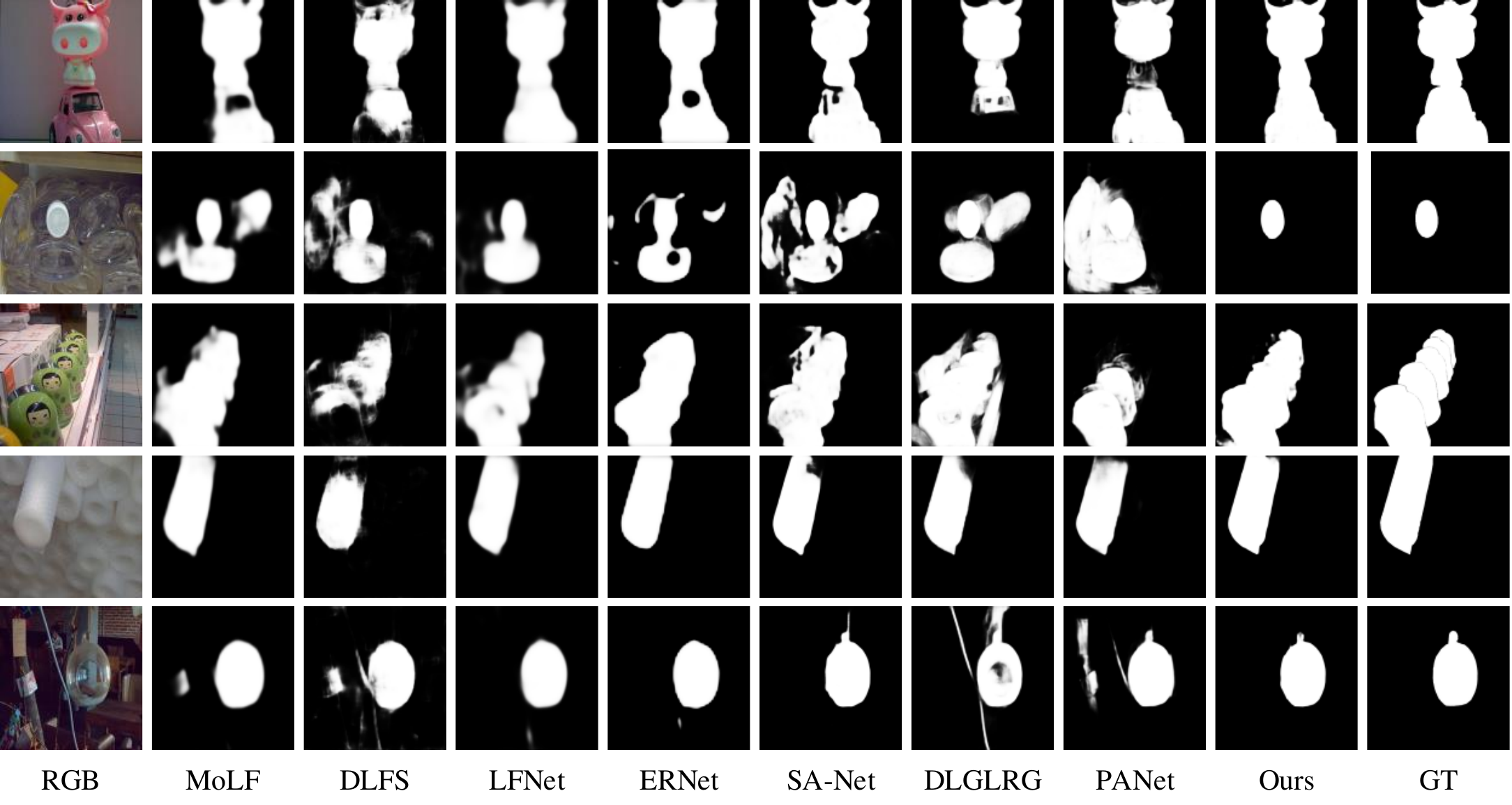}
	\caption{Visual comparison of LF SOD. Our HRTransNet is outstanding in some challenging cases: big objects ($1^{st}$ row), small objects ($2^{nd}$ row), multiple objects ($3^{rd}$ row),  similar foreground and background ($4^{th}$ row), and complex scenes ($5^{th}$ row).}
\label{fig:LFTvisual_compare}
\end{figure*}

\begin{table*}[!htp]
  \centering
  \fontsize{6}{10}\selectfont
  \renewcommand{\arraystretch}{0.4}
  \renewcommand{\tabcolsep}{0.4mm}
  \scriptsize
  \caption{The comparison of S-measure, adaptive F-measure, adaptive E-measure, MAE of different LF SOD models.}
\label{tab:LFcomparison}
  \scalebox{1}{
  \begin{tabular}{c|c|ccccccc|c}
  \hline\toprule
    Datasets& Metric &
       MoLF\cite{zhang2019memory} &DLFS\cite{piao2019deep} &LFNet\cite{zhang2020lfnet}  &ERNet\cite{piao2020exploit} &SA-Net\cite{zhang2021SANet}&DLGLRG\cite{liu2021light} &PANet\cite{piao2021panet} &HRTransNet\\
       & &NeurIPS19 &IJCAI19 &TIP20  &AAAI20&BMVC21 &ICCV21 &TCYB21 &Ours  \\

  \midrule

  \multirow{3}{*}{\textit{LFSD}}
    & $S\uparrow$    & .830 & .735 & .806    &.835&.841 &.867&.849&$\textbf{.875}$\\

    & $F_\beta \uparrow$    &.819 &.713 &.793 &.839 &.845&.861 &.849 &\textbf{.875}     \\

    & $E_{\xi}\uparrow$       &.886 & .805 & .870  &.887 &.889&\textbf{.898}&.893 &{.896}\\

    & MAE$\downarrow$         &.089 &.149 &.101 &.082 &.074&.069&.076 &\textbf{.056}\\
    \midrule

  \multirow{3}{*}{\textit{HFUT-Lytro}}
    & $S\uparrow$      &.742 &.741 &.781 &.778 &.784&.766&.795 &$\textbf{.827}$ \\

    &$F_\beta$$\uparrow$     &.627 &.616 &.659 &.705 &.736&.709&.724& $\textbf{.774}$\\

    & $E_{\xi}\uparrow$       & .785 &.784 & .808     &.831 &.850&.841 &.851&\textbf{.869}  \\

    & MAE$\downarrow$          & .095 &.097 & .076 &.082&.078 &.071&.074&$\textbf{.062}$ \\

    \midrule

  \multirow{3}{*}{\textit{DUTLF-FS}}
    & $S\uparrow$     &.887 &.841 &.882 &.900&.918 &.928&.908&$\textbf{.938}$\\

    & $F_\beta$$\uparrow$    &.843 &.801 &.842 &.888&.920&.923 &.897&\textbf{.944}\\

    & $E_{\xi}\uparrow$     & .923 & .891 & .914  &.942&.954&.952 &.940& $\textbf{.962}$\\

    & MAE$\downarrow$       &.052 &.076 &.054 &.040&.032&.031&.039&$\textbf{.024}$\\

  \bottomrule
  \hline
  \end{tabular}}

\end{table*}
\textbf{Qualitative Evaluation.} The visual examples in Fig. \ref{fig:LFTvisual_compare} show the performance of our proposed model in some challenging cases: big objects ($1^{st}$ row), small objects ($2^{nd}$ row), multiple objects ($3^{rd}$ row),  similar foreground and background ($4^{th}$ row), and complex scenes ($5^{th}$ row).
It suggests that our method can model long-range dependency and detect salient objects from a global perspective, meanwhile maintaining fine-grained  detail of the local region.

\subsection{Ablation studies}
We take as an example RGB-D SOD to conduct the ablation study about the selection of primary stream backbone and auxiliary stream backbone, effectiveness of supplementary modality injection module and dual-direction short connection fusion module.
\subsubsection{Selection of primary stream backbone}\label{sec:primarystream}
Some backbones are selected as feature extractors for primary stream.
They are
ResNet\cite{he2016deep}, Swin Transformer\cite{liu2021swin}, HRNet\cite{sun2019deep}, HRFormer\cite{yuan2021hrformer}.
Table. \ref{tab:PrimaryBackboneAblation} shows the results which indicate that HRFormer plays an important role in improving the performance. Moreover, the model with HRFormer backbone has the less computation cost.
\begin{table*}[!htp]
  \centering
  \fontsize{8}{10}\selectfont
  \renewcommand{\arraystretch}{1}
  \renewcommand{\tabcolsep}{1mm}
  \scriptsize
  %\captionsetup{labelformat=empty}
  \caption{Ablation study about selection of primary modality backbone.}
\label{tab:PrimaryBackboneAblation}
  \scalebox{1}{
  \begin{tabular}{c|c|c|cccc|cccc|cccc|cccc}
  \hline\toprule

   \multirow{2}{*}{\centering Variant}&\multirow{2}{*}{\centering Params(M)$\downarrow$}&\multirow{2}{*}{\centering FLOPs(G)$\downarrow$}&\multicolumn{4}{c|}{\centering NLPR} & \multicolumn{4}{c|}{\centering NJU2K} & \multicolumn{4}{c|}{\centering STERE} & \multicolumn{4}{c}{\centering SIP}\\
      &\multicolumn{1}{c|}{}& \multicolumn{1}{c|}{}
      &$S\uparrow$ &$F_\beta\uparrow$ &$E_{\xi}\uparrow$&MAE$\downarrow$
   &$S\uparrow$ &$F_\beta\uparrow$ &$E_{\xi}\uparrow$&MAE$\downarrow$
    &$S\uparrow$ &$F_\beta\uparrow$ &$E_{\xi}\uparrow$&MAE$\downarrow$
     &$S\uparrow$ &$F_\beta\uparrow$ &$E_{\xi}\uparrow$&MAE$\downarrow$ \\
  \hline
  ResNet\cite{he2016deep}
    &67.7 &22.3
    &.909 &	.865 &	.947 &	.029 &	.882 &	.864 &	.903 &	.051 	& .892 &	.862 &	.913 &	.045 &	.849 &	.835 &	.908 &	.064\\
    Swin Transformer\cite{liu2021swin} &119.9 &23.7
    &.908 &.866& 	.949& 	.028& 	.899&	.879 &	.916& 	.041 &	.879 &	.841 &	.907& 	.049 &	.880& 	.872& 	.925 &.048\\
    HRNet\cite{sun2019deep}
    & 81.2&21.5
    &.932 &	.907 &	.961 &	.020 &	.922 &	.916 &	.928 &	.031 	&.909 	& .889 	&.926 &	.036 &	.902 &	.902 &	.935 	&.038    \\

    HRFormer\cite{yuan2021hrformer}&\textbf{58.9}&\textbf{17.1}
&\textbf{.942} &\textbf{.919} &\textbf{.969} &\textbf{.016} &\textbf{.933}
    &\textbf{.928} &\textbf{.931} &\textbf{.026} &\textbf{.921} &\textbf{.904} &\textbf{.930} &\textbf{.030} &\textbf{.909}
    &\textbf{.916} &\textbf{.943} &\textbf{.035}\\
    \bottomrule
    \hline
  \end{tabular}}

\end{table*}

\subsubsection{Selection of auxiliary stream backbone}\label{sec:auxiliarystream}
Some backbones are selected as feature extractors for auxiliary stream.
They are
ResNet18\cite{he2016deep}, ResNet50\cite{he2016deep}, HRFormer\cite{yuan2021hrformer}, Swin Transformer\cite{liu2021swin}, Segformer\cite{xie2021segformer}, ConvNet\cite{liu2022convnet}.
Table. \ref{tab:AuxiliaryBackboneAblation} shows the results which indicate that ResNet18 can achieve comparable performance with a fewer parameters and FLOPs. Therefore, the final model adopts ResNet18 as the auxiliary stream backbone.
\begin{table*}[!htp]
  \centering
  \fontsize{8}{10}\selectfont
  \renewcommand{\arraystretch}{1}
  \renewcommand{\tabcolsep}{1mm}
  \scriptsize
  %\captionsetup{labelformat=empty}
  \caption{Ablation study about selection of supplementary modality backbone.}
\label{tab:AuxiliaryBackboneAblation}
  \scalebox{1}{
  \begin{tabular}{c|c|c|cccc|cccc|cccc|cccc}
  \hline\toprule

   \multirow{2}{*}{\centering Variant}&\multirow{2}{*}{\centering Params(M)$\downarrow$}&\multirow{2}{*}{\centering FLOPs(G)$\downarrow$}&\multicolumn{4}{c|}{\centering NLPR} & \multicolumn{4}{c|}{\centering NJU2K} & \multicolumn{4}{c|}{\centering STERE} & \multicolumn{4}{c}{\centering SIP}\\
      &\multicolumn{1}{c|}{}& \multicolumn{1}{c|}{}
      &$S\uparrow$ &$F_\beta\uparrow$ &$E_{\xi}\uparrow$&MAE$\downarrow$
   &$S\uparrow$ &$F_\beta\uparrow$ &$E_{\xi}\uparrow$&MAE$\downarrow$
    &$S\uparrow$ &$F_\beta\uparrow$ &$E_{\xi}\uparrow$&MAE$\downarrow$
     &$S\uparrow$ &$F_\beta\uparrow$ &$E_{\xi}\uparrow$&MAE$\downarrow$ \\
  \hline
  ResNet18\cite{he2016deep}
    &\textbf{58.26} &\textbf{17.12}
    &\textbf{.942} &\textbf{.919} &\textbf{.969} &\textbf{.016}
    &\textbf{.933} &\textbf{.928} &.931 &.026
    &.921 &.904 &.930 &.030
    &.909 &.916 &\textbf{.943} &.035
    \\
    ResNet50\cite{he2016deep}
    &82.36 &22.07
    &\textbf{.942} &.915 &\textbf{.969} &\textbf{.016}
    &.930 &.923 &.930 &.027
    &.921 &.902 &\textbf{.932} &.030
    &\textbf{.910} &\textbf{.918} &\textbf{.943} &\textbf{.034}
    \\HRFormer\cite{yuan2021hrformer}&88.75&27.23
&.938&	.913&.965 &.017&
.931 &	.924 	&.930& 	.027& 	.921 &	.903 &\textbf{.932} &	.031&	.907 &	.915 &	.940 &	.035
    \\
    Swin Transformer-B\cite{liu2021swin} &137.63 &31.38
    &.940 &.916 &.967 &.017
    &.932 &.927 &\textbf{.933} &.026
    &\textbf{.923} &\textbf{.906} &\.931 &\textbf{.029}
    &.909 &.912 &.941 &\textbf{.034}
    \\
    Segformer-B4\cite{xie2021segformer}
    &108.04 &24.02
    &.940 &.917 &.966 &.017
    &\textbf{.933} &\textbf{.928} &\textbf{.933} &\textbf{.025}
    &.921 &.904 &\textbf{.932} &.030
    &\textbf{.910} &\textbf{.918} &\textbf{.943} &\textbf{.034}
    \\
    ConvNeXt-B\cite{liu2022convnet}
    &138.52 &31.62
    &.938 &.914 &.964 &.017
    &.928 &.921 &.931 &.029
    &.920 &.902 &.930 &.030
    &.909 &.915 &.941 &.035
    \\
    \bottomrule
    \hline
  \end{tabular}}

\end{table*}

\subsubsection{Effectiveness of modules}
\begin{figure}[htp!]
	\centering
\begin{tabular}{c}
  \includegraphics[width=0.9\linewidth]{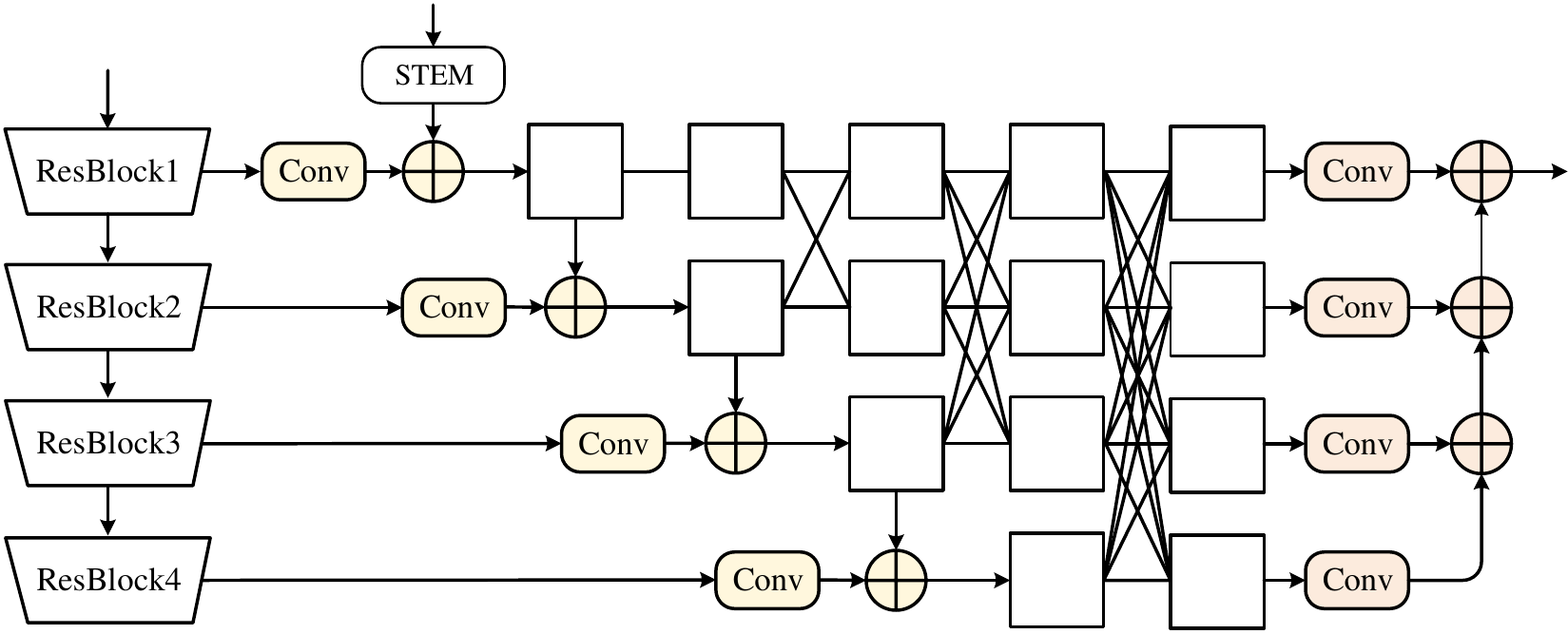}\\
  \small(a) Baseline \\
\includegraphics[width=0.9\linewidth]{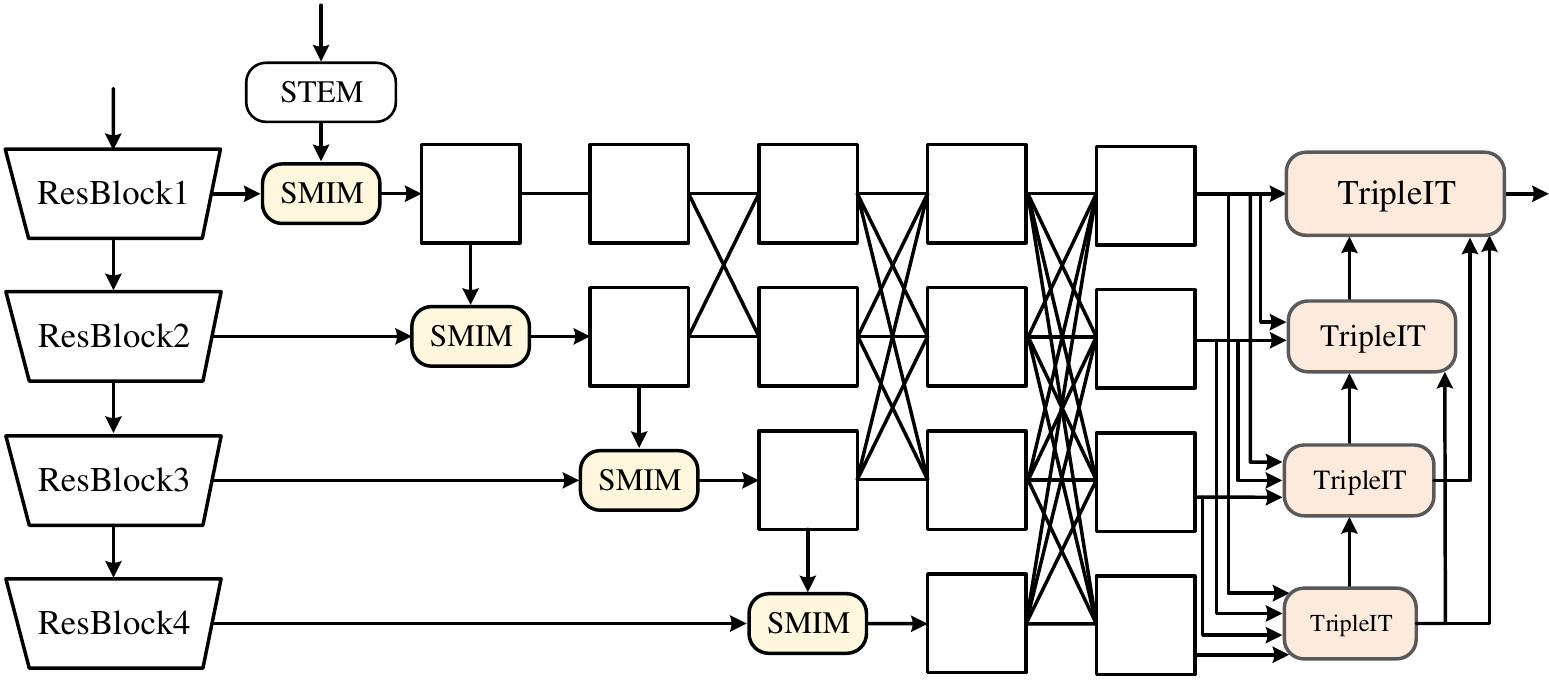}\\
\small(b) Ours
\end{tabular}
	
	\caption{The baseline and ours in the ablation study about SMIM and four TripleITs. }
     \label{fig:ResponseModuleEffectiveness}
\end{figure}

To offer deeper insights into supplementary modality injection module (SMIM) and the dual-direction short connection fusion module with four TripleITs, we perform the ablation study.
Fig. \ref{fig:ResponseModuleEffectiveness} shows the baseline model  and our model. The baseline model replaces SMIM with addition, and replaces four TripleITs with progressive decoding process with the addition  and upsampling.
%The ``Add" refers to the element-wise addition of two-modality features. The ``Ours" denotes our designed fusion strategy.

From Table. \ref{tab:ResponseModuleEffective}, we can see that all the evaluation values are improved except for S-measure and MAE in STERE dataset when adding SMIM to the baseline model. It illustrates that SMIM plays a positive role in improving the performance.  It benefits from the weighted fusion of two-modality data and the special coordinate attention assigned on auxiliary features with the latent noise.  Furthermore, TripleITs achieve the impressive improvement in all the evaluation metrics. It illustrates the obvious effect of four TripleITs by absorbing the advantage of  all the features with different resolutions to achieve complementary fusion and comprehensive detail enhancement. Finally, the model equipped with SMIM and TripleITs achieves the best performance.
    \begin{table*}[!htp]

  \centering
  \fontsize{8}{10}\selectfont
  \renewcommand{\arraystretch}{1}
  \renewcommand{\tabcolsep}{1mm}
  \scriptsize
  %\captionsetup{labelformat=empty}
  \caption{The ablation study about SMIM and four TripleITs.}
\label{tab:ResponseModuleEffective}
  \scalebox{1}{
\begin{tabular}{c|ccc|cccc|cccc|cccc|cccc}
    \hline
  \multirow{2}{*}{\centering Variant} &\multicolumn{3}{c|}{\centering Candidate} &\multicolumn{4}{c|}{\centering NLPR} & \multicolumn{4}{c|}{\centering NJU2K} & \multicolumn{4}{c|}{\centering STERE} &\multicolumn{4}{c}{\centering SIP}\\

  &Baseline&SMIM&TripleITs
  &$S\uparrow$ &$F_\beta\uparrow$ &$E_{\xi}\uparrow$&MAE$\downarrow$
   &$S\uparrow$ &$F_\beta\uparrow$ &$E_{\xi}\uparrow$&MAE$\downarrow$
    &$S\uparrow$ &$F_\beta\uparrow$ &$E_{\xi}\uparrow$&MAE$\downarrow$
     &$S\uparrow$ &$F_\beta\uparrow$ &$E_{\xi}\uparrow$&MAE$\downarrow$\\
     \hline

  No.1 &$\checkmark$&&&.928 &	.892 &	.956 	&.021 &	.925 &	.911 &	.921 &	.030 &	.917 &	.893 &	.924 &	.033 &	.899 &	.888 &	.930 &	.040\\

  No.2 &$\checkmark$&$\checkmark$&&.935 &.910 &.963 &.018 &.928 &.919 &.928 &.028 &.914 &.897 &.928 &.034 &.908 &.912
    &.940 &.036\\

    No.3 &$\checkmark$&&$\checkmark$&.936 &.909 &.963 &.019 &.932 &.926 &.930 &\textbf{.026} &.920 &.903 &\textbf{.932} &.031 &\textbf{.909} &.911 &.940 &\textbf{.035} \\

   No.4 &$\checkmark$&$\checkmark$&$\checkmark$&\textbf{.942} &\textbf{.919} &\textbf{.969} &\textbf{.016} &\textbf{.933}
    &\textbf{.928} &\textbf{.931} &\textbf{.026} &\textbf{.921} &\textbf{.904} &.930 &\textbf{.030} &\textbf{.909}
    &\textbf{.916} &\textbf{.943} &\textbf{.035} \\
    \hline
\end{tabular}}

\end{table*}
%\begin{table*}[!htp]
%  \centering
%  \fontsize{8}{10}\selectfont
%  \renewcommand{\arraystretch}{1}
%  \renewcommand{\tabcolsep}{1mm}
%  \scriptsize
%  %\captionsetup{labelformat=empty}
%  \caption{Ablation study of supplementary modality injection module. The best result is in bold.}
%\label{tab:SMIMAblation}
%  \scalebox{1}{
%\begin{tabular}{c|cccc|cccc|cccc|cccc}
%    \hline
%  \multirow{2}{*}{\centering Variant} &\multicolumn{4}{c|}{\centering NLPR} & \multicolumn{4}{c|}{\centering NJU2K} & \multicolumn{4}{c|}{\centering STERE} &\multicolumn{4}{c}{\centering SIP}\\
%  &$S\uparrow$ &$F_\beta\uparrow$ &$E_{\xi}\uparrow$&MAE$\downarrow$
%   &$S\uparrow$ &$F_\beta\uparrow$ &$E_{\xi}\uparrow$&MAE$\downarrow$
%    &$S\uparrow$ &$F_\beta\uparrow$ &$E_{\xi}\uparrow$&MAE$\downarrow$
%     &$S\uparrow$ &$F_\beta\uparrow$ &$E_{\xi}\uparrow$&MAE$\downarrow$
%     \\
%    \hline
%
%   Add&0.936 &0.909 &0.963 &0.019 &0.932 &0.926 &0.930 &\textbf{0.026} &0.920 &0.903 &\textbf{0.932} &0.031 &\textbf{0.909} &0.911 &0.940 &\textbf{0.035} \\
%
%     Ours&\textbf{0.942} &\textbf{0.919} &\textbf{0.969} &\textbf{0.016} &\textbf{0.933}
%    &\textbf{0.928} &\textbf{0.931} &\textbf{0.026} &\textbf{0.921} &\textbf{0.904} &0.930 &\textbf{0.030} &\textbf{0.909}
%    &\textbf{0.916} &\textbf{0.943} &\textbf{0.035} \\
%    \hline
%\end{tabular}}
%
%\end{table*}

\begin{figure}[!htp]
	\centering	\includegraphics[width=1\linewidth]{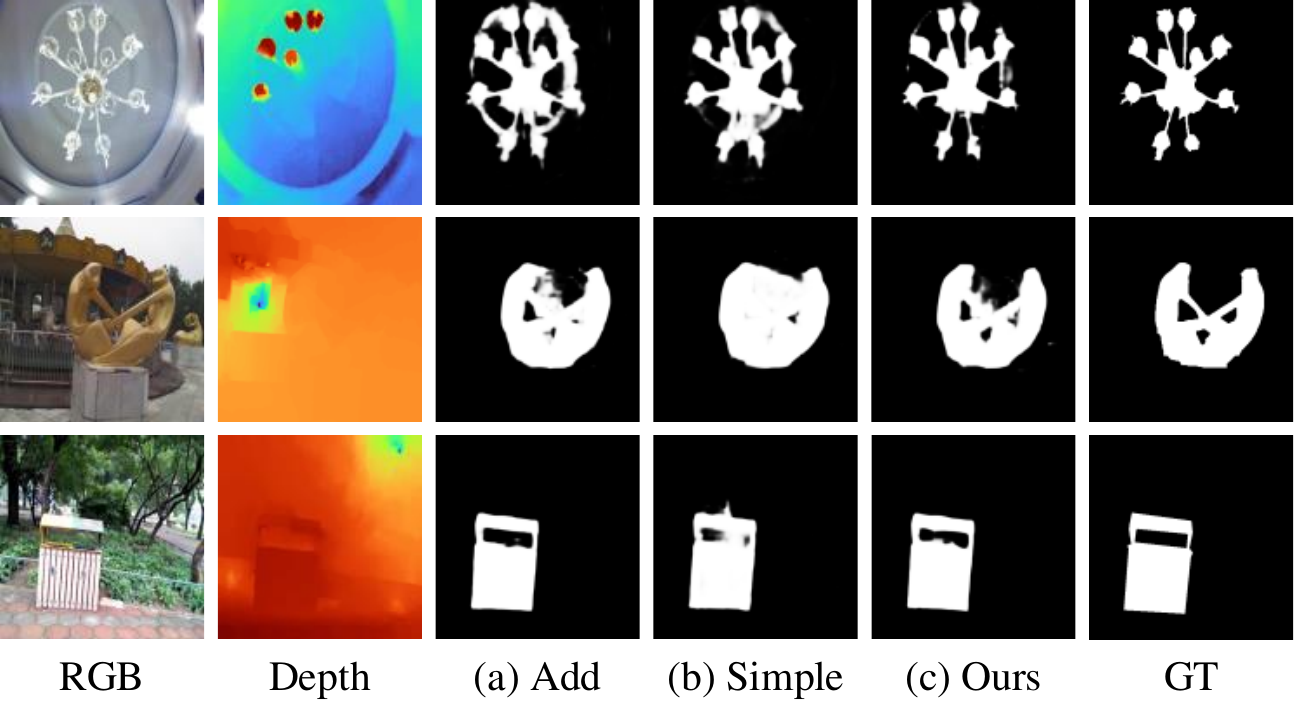}
	\caption{Visual comparison. From the comparison between (a) and (c), we can verify the effectiveness of supplementary modality injection module. From the comparison between (b) and (c), we can prove the advantage of dual-direction short connection fusion module.} \label{fig:VisualAblation}
\end{figure}

We also present  some visual comparisons in Fig. \ref{fig:VisualAblation}. From the comparison of (a) and (c), we discover that addition operation is inferior to our proposed method in the depth modality injection process, especially when depth modality is incomplete or inadequate, which is demonstrated by the first and second lines.
Compared with addition, ours can assign different weight to each modality and better suppress the noise in the depth modality. Accordingly the better results can be achieved.
From the comparison of (b) and (c), we find the simple decoder can locate the salient object accurately but generate blurry boundaries. It indicates the simple decoder model has no advantage in local detail representation. Compared with it, our model is better in polishing details by the long-range dependency of the transformer and achieving  excellent performance with less noise.

\subsection{Failure cases}
\begin{figure*}[!htp]
	\centering	\includegraphics[width=1\textwidth]{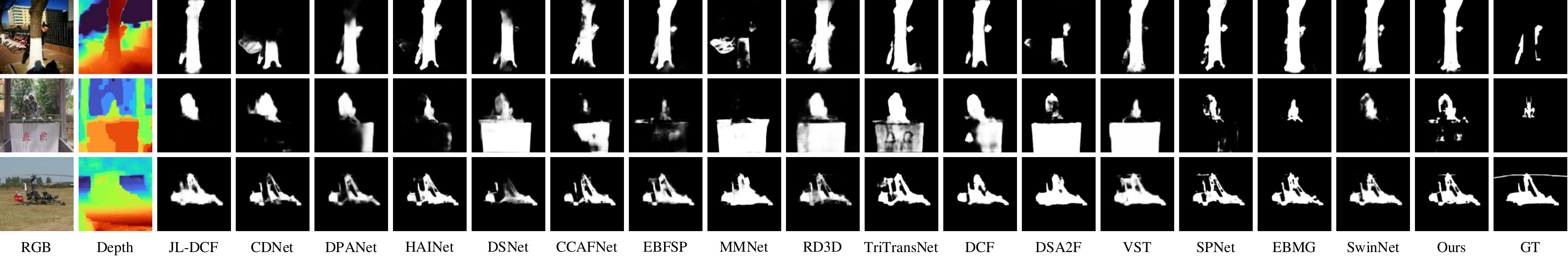}
	\caption{Failure cases. RGB-D SOD methods all fail in occlusion (1$^{st}$ line), extreme background interference (2$^{nd}$ line), and object completeness (3$^{rd}$ line).} \label{fig:Failurecase}
\end{figure*}

The proposed model has a good detection performance in most
cases. However, there are a few failure cases, as
shown in Fig. \ref{fig:Failurecase}. The first line can't highlight the  person due to serious occlusions. The second line generates the error result due to extremely similar background. The propeller in third line is incomplete because it is misidentified as the background instead of the part of the glider. The similar failed detection results also occur in  the other state-of-the-art RGB-D SOD methods.
Therefore, salient object detection need to be further studied to solve the occlusions, extreme background interference, and object completeness problems.

\section{Conclusions}
In this paper, we study two-modality salient object detection tasks based on HRFormer.
For the one-modality to two-modality extension, we inject a
supplementary modality to effectively combine two-modality cues.
We improve HRFormer by modifying the final multi-resolution fusion using intra-feature and inter-feature interactive transformers. Each transformer-like unit enhances the feature by the feature itself and the features with  other resolutions. The transmission of features to TripleIT constitutes a dual-direction short connection flow.
The proposed HRTransNet is applied to RGB-D, RGB-T, and LF SOD, and exhibits excellent performance. We will attempt other two-modality SOD tasks and exploit multi-modality SOD and co-saliency tasks in future studies. Furthermore, we will discuss SOD task in virtual reality and augmented reality (VR/AR) application.

%\section*{Acknowledgment}


%\bibliographystyle{IEEEtran}
%\bibliography{HRTransNetbib}
%
%
%\vfill

%\clearpage
% \onecolumn

%\begin{figure}[!htp]
%	\includegraphics[width=0.1\linewidth]{Bin_Tang.jpg}
%\label{fig:Bin_Tang}
%\end{figure}
%\textbf{Bin Tang} is a lecturer in School of Artificial Intelligence and Big Data, Hefei University, China. He received his Ph.D. from Fudan University, China in 2008. His research interests include computer vision.
%
%
%\begin{figure}[!htp]
%	\includegraphics[width=0.1\linewidth]{Zhengyi_Liu.jpg}
%\label{fig:Zhengyi_Liu}
%\end{figure}
%\textbf{Zhengyi Liu} is a professor in School of Computer Science and Technology, Anhui University, China. She received her B.S., M.S., and Ph.D. from Anhui University, China in 2001, 2004 and 2007, respectively. Her research interests include image and video processing, computer vision.
%
%
%\begin{figure}[!htp]
%	\includegraphics[width=0.1\linewidth]{Yacheng_Tan.jpg}
%\label{fig:Yacheng_Tan}
%\end{figure}
%\textbf{Yacheng Tan} is a M.S. Candidate of Anhui University. He received his B.S. from Anhui Polytechnic University, China in 2020. His research interests include image and video processing, computer vision.
%
%\begin{figure}[!htp]
%	\includegraphics[width=0.1\linewidth]{Qian_He.jpg}
%\label{fig:Qian_He}
%\end{figure}
%\textbf{Qian He} is a M.S. Candidate of Anhui University. She received her B.S. from Jianghuai College of Anhui University, China in 2021. Her research interests include image and video processing, computer vision.

\end{document}